\documentclass{article}

\PassOptionsToPackage{square,comma,numbers,sort&compress}{natbib}
 \usepackage[preprint]{neurips_2026}

\usepackage[utf8]{inputenc} 
\usepackage[T1]{fontenc}    
\usepackage[pagebackref,breaklinks,colorlinks,citecolor=cvprblue]{hyperref}
\usepackage{url}            
\usepackage{booktabs}       
\usepackage{amsfonts}       
\usepackage{nicefrac}       
\usepackage{microtype}      
\usepackage{xcolor}         
\definecolor{cvprblue}{rgb}{0.21,0.49,0.74}

\usepackage{booktabs}
\usepackage{amsfonts}
\usepackage{amsmath}
\usepackage{amsthm}
\usepackage{amssymb}
\usepackage{mathtools}
\usepackage{bm}
\usepackage{nicefrac}
\usepackage{xcolor}
\usepackage{float}
\usepackage{colortbl}
\usepackage{graphicx}
\usepackage{wrapfig}
\definecolor{mygray}{gray}{0.9}
\usepackage{enumitem}
\usepackage{multirow}
\usepackage{subcaption}
\usepackage{titletoc}

\theoremstyle{plain}

\newtheorem*{definition*}{Definition}

\newtheorem*{assumption*}{Assumption}
\newtheorem*{principle*}{Principle}

\newtheorem*{theorem*}{Theorem}
\newtheorem*{proposition*}{Proposition}

\newtheorem*{corollary*}{Corollary}

\newtheorem*{lemma*}{Lemma}

\usepackage{algorithm}
\usepackage{algpseudocode}

\definecolor{highlightred}{gray}{0.9}
\usepackage{pifont}

\title{Latent Thought Flow: Efficient Latent Reasoning in Large Language Models}

\author{%
  Xiandong Zou$^{1}$, Jing Huang$^{2}$, Jianshu Li$^{2}$, Pan Zhou$^{1}$\thanks{Corresponding author.} \\
  $^{1}$Singapore Management University $^{2}$Ant Group\\
}

\begin{document}

\maketitle
\begin{abstract}
Large Language Models (LLMs) increasingly rely on intermediate reasoning, yet explicit Chain-of-Thought (CoT) suffers from a linguistic space bottleneck: each thought must be decoded into tokens, causing high inference overhead. Latent reasoning moves deliberation into continuous space, but existing methods mostly learn deterministic or reward-maximizing paths, lacking a principled way to allocate probability across trajectories with different correctness and costs. We propose \textbf{Latent Thought Flow (LTF)}, which models reasoning as variable-length continuous trajectories and trains a sampler to match a reward-induced posterior over answer quality and computation cost. We instantiate this with a continuous GFlowNet using stochastic latent transitions. To handle sparse answer supervision, we introduce an Entropy-Weighted Subtrajectory Balance objective for intermediate rewards and a reference-prior regularizer to anchor exploration. Experiments under finetuning and transfer learning settings show that LTF outperforms explicit CoT and latent reasoning baselines, improving accuracy by 9.5\% while reducing reasoning length by 27.2\% on average compared with strong latent reasoning baselines.
\end{abstract}

\section{Introduction}
\label{sec:introduction}
Large Language Models (LLMs)~\citep{gpt4,gpt5,llama3,gemini} are increasingly tasked with complex objectives that demand a synergy of perception and deliberation, ranging from math problem solving~\citep{lightman2023let,gao2023pal}, embodied planning~\citep{shinn2023reflexion,hao2023reasoning}, and code generation~\citep{li2022competition,nijkamp2022codegen}. A cornerstone of this success is the use of intermediate reasoning processes, where frameworks like Chain-of-Thought (CoT) \citep{cot, kojima2022large}, Self-Consistency \citep{wang2023selfconsistency}, and Tree-of-Thoughts \citep{yao2023tree} decompose intricate problems into sequential rationales. While promising, they inevitably encounter a \textit{linguistic space bottleneck}: the requirement to decode every intermediate thought into discrete tokens. The explicit reasoning not only incurs substantial inference overhead but also constrains the model’s internal computation into a potentially redundant textual form. Consequently, this reliance on explicit rationales creates an inherent accuracy--efficiency trade-off, limiting the scalability of reasoning capabilities in resource-constrained scenarios.

Latent reasoning has emerged as a promising alternative to mitigate the linguistic space bottleneck by performing multi-step computation directly within the model's continuous hidden-state space. Rather than fully externalizing deliberation as discrete text, these approaches keep intermediate reasoning internalized: Pause tokens \citep{goyal2024pause} allocate additional latent computation; stepwise internalization and compressed CoT \citep{icot, cheng2024compressed} densify textual rationales; and other frameworks~\citep{coconut, codi, softcot,softcot2,jiangrethinking,yue2025hybrid,zhaolearning} like Coconut propagate continuous thought vectors or soft representations via the LLM backbone. These works demonstrate that meaningful intermediate computation need not always be expressed in linguistic words. However, most existing approaches primarily define a deterministic latent reasoning path and train it by compression, distillation, or reward maximization under a fixed reasoning budget. They do not specify how sampling probability mass should be allocated across latent reasoning trajectories with different correctness and computational costs under various questions.

We argue that this missing distributional view is central to efficient latent reasoning. For a given question, an LLM must navigate a manifold of plausible reasoning paths and only a sparse subset of which yields both logical consistency and computational efficiency. Without this distributional perspective, existing paradigms often fail to optimize the latent reasoning manifold: maximum-likelihood objectives inherit the verbosity of training rationales, while reinforcement learning tends to trigger posterior collapse onto a few high-reward modes. What is required is \textit{a principled sampler capable of modeling the trade-offs between correctness and computational cost.} Such a framework should concentrate probability mass on concise and accurate trajectories while suppressing redundant or erroneous paths, thereby enabling the model to dynamically allocate its computational budget.

In this paper, we propose \textbf{Latent Thought Flow (LTF)}, a reward-proportional framework for learning continuous latent reasoning in LLMs. Given a question $x$, LTF samples a latent reasoning trajectory $\tau=(\mathbf{z}_{1:T},\bot)$, where each $\mathbf{z}_t$ is a continuous thought state in LLM's latent space and $\bot$ denotes an adaptive stopping decision. The LLM then produces the final response conditioned on $x$ and $\tau$ (Fig.~\ref{fig:o}). During training, the target answer defines a utility $\mathcal{R}_{x,y}(\tau)$ consisting of answer quality and computation cost, inducing the desired posterior over latent thoughts,
\begin{equation}
	p^*(\tau\mid x,y) \propto \mathcal{R}_{x,y}(\tau).
\end{equation}
We train the latent sampler to match this reward-induced distribution with a continuous GFlowNet objective. GFlowNets are designed to learn stochastic generative policies whose terminal samples are proportional to an unnormalized reward, which makes them well suited for preserving diverse high-utility solutions rather than only optimizing for a single maximum-reward trajectory~\citep{gflownet,gflownet1}. While recent work has applied GFlowNets to LLM or VLM reasoning over discrete texts, proof, or action traces~\citep{hu2024amortizing,takase2024gflownet,kang2025gflowvlm}, LTF moves reward-proportional trajectory learning into LLM's latent space.

This formulation introduces three technical challenges. First, latent thoughts are high-dimensional continuous vectors, so discrete token-level balance equations are not directly applicable. Building on continuous GFlowNet theory~\citep{cgflow}, LTF parameterizes forward transitions with conditional densities and writes balance constraints over variable-length continuous trajectories. Second, answer-level supervision is sparse: the utility of a latent subtrajectory is typically observed only after the final response. We therefore use a Subtrajectory Balance objective~\citep{tb,subtb} with entropy-aware weighting to propagate terminal rewards to intermediate latent subtrajectories. Third, unconstrained exploration in a continuous latent space can drift away from meaningful latent states. LTF addresses this with a reference-prior regularizer that warm-starts exploration from modality-aligned anchors, while annealing the prior so that learned trajectories are shaped by the accuracy--efficiency reward. During inference, LLM samples latent thought paths and generates answers without access to gold rationales.

Experiments demonstrate that LTF improves the accuracy--efficiency frontier of reasoning, reduces reasoning overhead, and adapts its latent budget to problem difficulty compared with explicit CoT, latent reasoning, and reward-maximizing baselines. Compared with strong latent reasoning baselines CoLaR and ReGuLaR, LTF improves accuracy by 12.9\% while reducing reasoning length by 34.5\% on average in fine-tuning tasks; under transfer learning, LTF improves accuracy by 6.0\% while reducing reasoning length by 19.9\% on average.

\section{Related Work}
\label{sec:related_work}

\paragraph{Explicit Chain of Thought (CoT).}
A common approach to reasoning exposes intermediate computation as text. CoT prompting elicits rationales~\citep{cot,kojima2022large}, self-consistency aggregates sampled rationales~\citep{wang2023selfconsistency}, and Tree-of-Thoughts searches textual states~\citep{yao2023tree}. Multimodal variants produce explanations or grounded rationales, e.g., ScienceQA~\citep{lu2022learn}, Multimodal-CoT~\citep{zhang2024multimodal}, and Visual CoT~\citep{shao2024visualcot}, while tool- and program-based systems externalize reasoning via vision experts or executable programs~\citep{yang2023mmreact,suris2023viper}. Although interpretable, these methods serialize visual, spatial, and perceptual computation into lengthy language, programs, or actions. LTF instead uses latent trajectories, reducing decoding overhead while retaining adaptive computation before final answer generation.

\paragraph{Latent Reasoning.}
To reduce the cost of long textual rationales, recent work explores reasoning in LLM latent spaces. Pause tokens add hidden computation before decoding~\citep{goyal2024pause}; explicit-to-implicit training gradually removes CoT supervision~\citep{icot}; and continuous-thought methods replace rationales with dense contemplation tokens, recurrent states, or soft thought tokens~\citep{cheng2024compressed,coconut,codi,softcot,softcot2,sun2026llm}. Other methods learn abstract discrete tokens~\citep{ramji2026abstract} or probabilistic latent vectors with posterior inference~\citep{kong2025latent,jiangrethinking,colar,regular}. Unlike methods that imitate or compress a single reasoning trace, LTF learns a reward-induced posterior over variable-length latent reasoning trajectories.

\paragraph{GFlowNets for Diverse Reasoning.}
Generative Flow Networks learn a policy whose terminal distribution is proportional to an unnormalized reward~\citep{gflownet,gflownet1}, providing a distribution-matching alternative to reinforcement-learning-based latent reasoning that typically maximizes expected reward~\citep{zhaolearning,du2025latent,yue2025hybrid,zhou2026lepo}. By preserving probability mass over multiple high-reward solutions, GFlowNets naturally support diverse reasoning. Trajectory Balance~\citep{tb} and continuous-space GFlowNets~\citep{cgflow} further enable textual reasoning~\citep{gthink,zhu2025flowrl} and visual rationale generation~\citep{hu2024amortizing,kang2025gflowvlm,gthink1}. However, existing GFlowNet-based methods mainly model explicit textual, symbolic, or visual traces, with limited focus on reasoning efficiency. In contrast, LTF samples in continuous latent superposition space, enabling adaptive computation and favoring trajectories that are both accurate and efficient.


\vspace{-0.5em}
\section{Methodology}
\label{sec:method}
\vspace{-0.5em}
\paragraph{Overview.}
\begin{figure}[t!]
  \centering
  \includegraphics[width=0.99\textwidth]{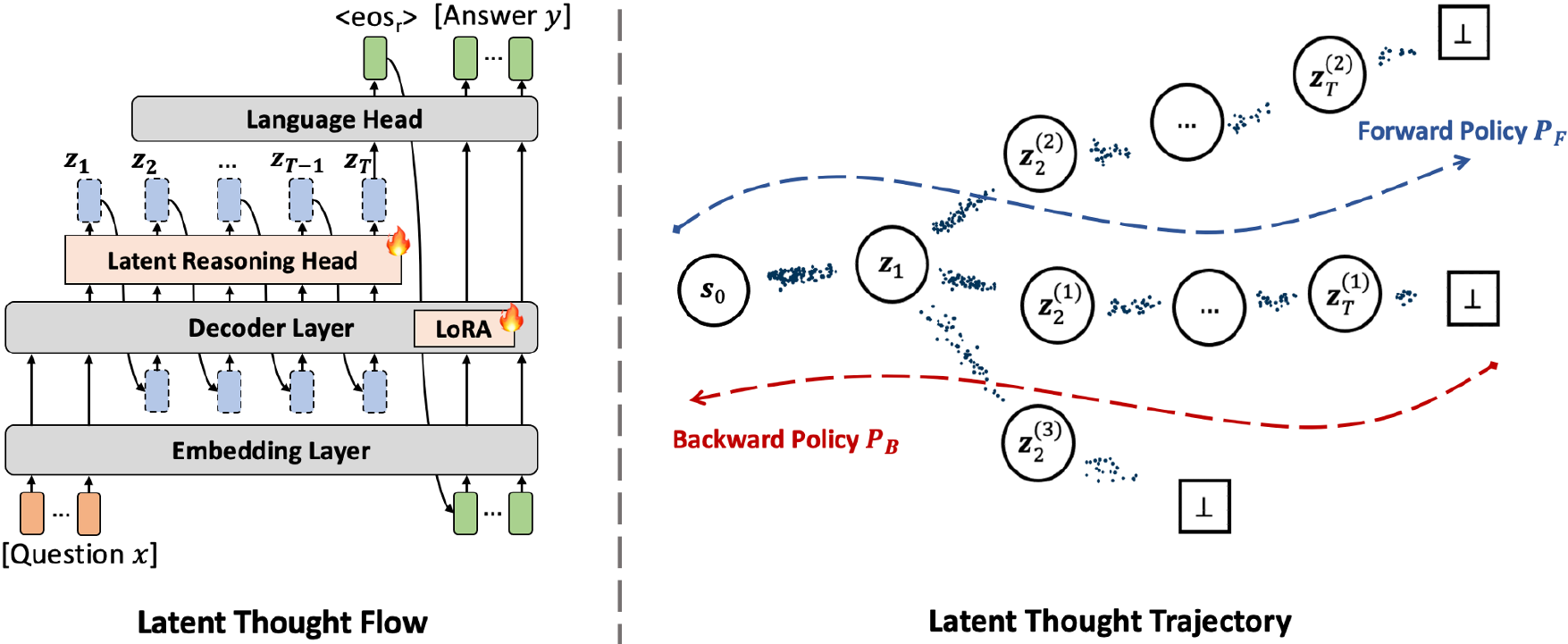}
  \caption{Illustration of our \textbf{LTF} framework. \textit{\textbf{Left:}} Overall architecture of \textbf{LTF}. Only the LoRA module and latent reasoning head are trainable. \textit{\textbf{Right:}} Overview of latent thought trajectories from a continuous GFlowNet perspective. A latent reasoning trajectory is represented as $\tau=(s_0,\mathbf{z}_1,\ldots,\mathbf{z}_T,\bot)$, where $\mathbf{z}_t$ denotes the latent thought state at step $t$. The forward policy $P_F$ is parameterized as a variational sampling via LLM with a latent head (Eq.~\eqref{eq:sample}) and trained with our entropy-weighted subtrajectory balance objective (Eq.~\eqref{eq:entropy_normalized_subtb}) to achieve the flow balance condition.}
\label{fig:o}
\vspace{-1em}
\end{figure}

Latent Thought Flow (LTF) models reasoning as an adaptive flow over continuous latent thought states before generating the final answer (Fig.~\ref{fig:o}). Rather than committing to a single deterministic latent thought chain or relying on explicit textual rationales, LTF learns a distribution over variable-length latent reasoning trajectories. Its guiding principle is that \textit{a trajectory should be sampled with high probability when it solves the problem accurately using minimal computation, and with low probability when it is incorrect or redundant.} We instantiate this principle with a continuous GFlowNet objective, which trains a reward-proportional sampler over latent trajectories~\citep{gflownet,gflownet1,cgflow}. Concretely, LTF consists of four components: a variable-length latent thought sampler (Sec.~\ref{sec:latent_trajectory}), an accuracy--efficiency reward (Sec.~\ref{sec:reward_target}), a continuous Subtrajectory Balance objective (Sec.~\ref{sec:continuous_gfn}), and a reference-prior regularizer (Sec.~\ref{sec:reference_prior}) that keeps early exploration semantically grounded.

\subsection{Variable-Length Latent Thought Trajectories}
\label{sec:latent_trajectory}

LTF replaces explicit textual rationales with an adaptive latent reasoning process: instead of forcing the model to learn deterministic reasoning tokens, it learns ``how much'' to reason internally before producing the final answer. Given a training set $\mathcal{D}=\{(x,y,r)\}$, where $x$ can be an input containing a question or instruction, $y$ is the target answer, and $r$ is an optional reference rationale used only for training, we would like to train an LLM $p_{\Theta}$ with the latent reasoning mechanism. Let $\Theta=\{\phi,\psi\}$ denote the model parameters, where $p_{\phi}$
consists of the embedding and decoder layers that produce contextual
latent states, and $p_{\psi}$ denotes the language head that generates explicit tokens (Fig.~\ref{fig:o}).

Formally, given an input $x$, LTF samples a variable-length latent thought trajectory
\begin{equation}
\tau=(\mathbf z_{1:T},\bot)\in\mathcal{T},\qquad \mathbf z_t\in\mathbb R^{d_z},\qquad 0\le T\le T_{\max},
\end{equation}
where $T$ denotes the number of continuous latent thoughts and $\bot$ denotes termination ($T=0$ means answering without reasoning). Let $h_x=p_\phi(x)$ and $s_t=(h_x,\mathbf z_{1:t})$. To build a variational latent reasoning process with expressive latent superposition states, each latent thought $\mathbf{z}_t$ is sampled by a sampler $q_\varphi$ consisting of a decoder layer and a latent reasoning head (Fig.~\ref{fig:o}).

At each step $t$, conditioned on the current context $s_t$, with $s_0=h_x$, the LLM samples each latent reasoning state from its posterior distribution given the question and the previous ones, i.e., $z_t\sim q_{\varphi}(\cdot| s_t).$
In this study, we model the latent policy $q_\varphi$ as the Gaussian distribution following prior work~\citep{colar,regular}, and it samples the latent thoughts from a continuous density iteratively until achieving the decoded $<\mathrm{eos_{r}}>$ token:
\begin{equation}
\label{eq:sample}
q_\varphi(\mathbf z_{t+1}\mid s_t)
=
\mathcal N\!\left(
\boldsymbol\mu_\varphi(s_t),
\operatorname{diag}(\boldsymbol\sigma_\varphi^2(s_t))
\right),
\qquad
\pi^\bot(s_t)=p_{\psi}(<\mathrm{eos_{r}}>|s_t),
\end{equation}
with the reparameterization $\mathbf z_{t+1}=\boldsymbol\mu_\varphi(s_t)+\boldsymbol\sigma_\varphi(s_t)\odot\boldsymbol\epsilon$, $\boldsymbol\epsilon\sim\mathcal N(0,I)$, which allows gradients from the answer loss to pass through the latent computation~\citep{kingma2014autoencoding}. The resulting trajectory density is
\begin{equation}
\label{eq:trajectory_density_ltf}
q_{\varphi}(\tau\mid x)
=
\left[
\prod_{t=0}^{T-1}
\bigl(1-\pi^\bot(s_t)\bigr)
q_\varphi(\mathbf z_{t+1}\mid s_t)
\right]
\pi^\bot(s_T),
\qquad
\tau=(\mathbf z_{1:T},\bot).
\end{equation}
where the product is empty when $T=0$. We force $\pi^\bot(s_{T_{\max}})=1$ when the maximum budget $T_{\max}$ is reached. Eq.~\eqref{eq:trajectory_density_ltf} provides the prerequisite for an adaptive reasoning budget by modeling a distribution over variable-length latent trajectories. This allows LTF, in principle, to assign shorter trajectories to easier examples and longer trajectories to harder ones.

Given the prefix and latent thought tokens, the answer can be generated:
\begin{equation}
p_\psi(y\mid x,\tau)
=
\prod\nolimits_{m=1}^{|y|}
p_\psi\!\left(
y_m\mid \tau,y_{<m}
\right).
\end{equation}
Learning such latent reasoning trajectories is challenging for three reasons. First, reasoning states are unobserved, continuous, and variable-length, with no token-level supervision for identifying informative intermediate computations. The model must infer both what to compute and when to stop from the final answer quality alone. Second, many trajectories may yield the same correct answer, but differ in computational efficiency, so optimizing a single best path is insufficient; probability should reflect both accuracy and cost. Third, latent thought spans a high-dimensional continuous space, where naive exploration is unstable and prone to semantically meaningless states.

\subsection{Reward-Proportional Target Distribution}
\label{sec:reward_target}

Since latent thoughts are unobserved, LTF specifies their desired behavior through a terminal reward rather than step-level supervision. For each training pair $(x,y)$, we assign a trajectory $\tau=(s_{T}\rightarrow\bot)$ a non-negative accuracy--efficiency utility 
\begin{equation}
	\label{eq:accuracy_efficiency_reward}
	\mathcal{R}_{x,y}(\tau)
	=
	V_{x,y}(\tau)
	\exp\bigl(-\lambda_c C(\tau)\bigr),
\end{equation}
where $V_{x,y}(\tau)$ measures answer quality, $C(\tau)$ measures latent computation, $\lambda_c$ controls the cost penalty. We instantiate the quality score as
\begin{equation}
	\label{eq:quality_score}
	V_{x,y}(\tau)
	=
	\operatorname{Ver}(y,\hat y_\tau)
	+
	\exp\left(
1/|y| \cdot	\log p_\psi(y\mid x,\tau) 
	\right),
\end{equation}
where $\hat y_\tau$ is the decoded answer and $\operatorname{Ver}$ denotes the task-specific accuracy. The verifier term aligns the reward with the evaluation metric, while the normalized likelihood term provides a dense signal. By default, we use a length-based cost function
\begin{equation}
	\label{eq:cost_proxy}
	C(\tau)=T,
\end{equation}
which favors shorter latent trajectories unless additional computation improves answer quality.

This reward induces the target distribution over terminal latent trajectories:
\begin{equation}
	\label{eq:reward_induced_distribution}
	p^*(\tau\mid x,y)
	= {\mathcal{R}_{x,y}(\tau)}/{Z_{\mathcal{R}}(x,y)},
	\qquad
	Z_{\mathcal{R}}(x,y)
	=
	\int_{\mathcal{T}}
	\mathcal{R}_{x,y}(\tau)
	d\mathbf \tau.
\end{equation}
LTF learns not a single optimal path, but a reward-proportional distribution where multiple correct and efficient trajectories coexist, while incorrect or inefficient ones receive low mass. Because $Z_{\mathcal{R}}(x,y)$ integrates over continuous latent trajectories of all lengths, direct normalization is intractable. We thus train the amortized sampler $q_\varphi(\tau\mid x)$ to match this target. The answer $y$ defines rewards only during training; at inference, the sampler conditions solely on $x$.

\subsection{Continuous GFlowNet}
\label{sec:continuous_gfn}

GFlowNets train a stochastic generator to sample complete trajectories in proportion to a positive reward, rather than only maximizing the expected reward of a single trajectory~\citep{gflownet,gflownet1}. This property is well suited for latent reasoning, where multiple internal thought paths may lead to the same correct answer and should therefore be preserved during training. In LTF, the latent thoughts are continuous vectors, so the discrete transition probability in standard GFlowNets becomes a continuous transition density. To provide dense supervision for intermediate latent prefixes, we allow every prefix state to terminate: after stopping at a prefix, the model decodes an answer and receives a reward. This allows us to compute the flow of each prefix analytically from its immediate-stop reward and stop probability, avoiding an additional learned flow estimator.

For a subtrajectory $s_i\to s_{i+1}\to\cdots\to s_j$, the GFlowNet flow balance condition requires
\begin{equation}
\label{eq:continuous_flow_consistency}
    F(s_i)
    \prod\nolimits_{t=i}^{j-1}
    P_F(s_{t+1}\mid s_t)
    =
    F(s_j)
    \prod\nolimits_{t=i}^{j-1}
    P_B(s_t\mid s_{t+1}),
\end{equation}
where $F(s_i)$ is the flow at state $s_i$, and $P_F$ and $P_B$ denote the forward and backward transitions, respectively. Since our state $s_t=(h_x,\mathbf z_{1:t})$ stores the full latent prefix, each state has a unique parent obtained by removing the last latent vector. Thus, the backward transition is deterministic and contributes zero log-density along a valid trajectory.

Under our latent thought trajectory formulation in Sec.~\ref{sec:latent_trajectory},
the forward transition density from $s_t$ to non-terminal state $s_{t+1}$ is given by
\begin{equation}
\label{eq:continuous_forward_density}
    P_F(s_{t+1}\mid s_t)
    =
    \bigl(1-\pi^\bot(s_t)\bigr)
    q_\varphi(\mathbf z_{t+1}\mid s_t),
\end{equation}
where $q_\varphi(\mathbf z_{t+1}\mid s_t)$ is the Gaussian density defined in Eq.~\eqref{eq:trajectory_density_ltf}, and $\pi^\bot(s_t)$ is the probability of stopping at $s_t$. Therefore, the forward edge log-density for a non-terminal transition is
\begin{equation}
\label{eq:continuous_forward_edge}
    \ell_t^{\varphi}
    =
    \log q_\varphi(\mathbf z_{t+1}\mid s_t)
    +
    \log\bigl(1-\pi^\bot(s_t)\bigr).
\end{equation}
Since the reward $\mathcal R_{x,y}(s_t\to\bot)$ of terminating at prefix $s_t$ is defined in Eq.~\eqref{eq:accuracy_efficiency_reward}, where the model stops at $s_t$, decodes an answer, and scores it against the training answer $y$, the flow $F(s_{t})$ can be expressed by the terminal consistency condition
\begin{equation}
\label{eq:prefix_terminal_flow}
    F(s_t)\pi^\bot(s_t)
    =
    \mathcal R_{x,y}(s_t \rightarrow \bot) \Rightarrow F(s_t)
    =
    \frac{
        \mathcal R_{x,y}(s_t \rightarrow \bot)
    }{
        \pi^\bot(s_t)
    }.
\end{equation}
Thus, the flow of an intermediate prefix does not need to be learned by a separate network; it can be computed from the reward obtained by immediately stopping at that prefix.

We use Sub-Trajectory Balance~\citep{subtb} objective to enforce this consistency over all subtrajectories due to its stable optimization and low variance property~\citep{ag,gflownet1}. For $0\le i<j\le T$, the residual is
\begin{equation}
\label{eq:continuous_subtb_residual}
\begin{aligned}
    \chi_{i:j}
    &=
    \log F(s_i)
    +
    \sum\nolimits_{t=i}^{j-1}\ell_t^{\varphi}
    -
    \log F(s_j) \\
    &=
    \log
    \frac{
        \mathcal R_{x,y}(s_i \rightarrow \bot)
        \left[
        \prod\nolimits_{t=i}^{j-1}
        q_\varphi(\mathbf z_{t+1}\mid s_t)
        \bigl(1-\pi^\bot(s_t)\bigr)
        \right]
        \pi^\bot(s_j)
    }{
        \mathcal R_{x,y}(s_j \rightarrow \bot)
        \pi^\bot(s_i)
    } .
\end{aligned}
\end{equation}
Thus, the continuous Sub-Trajectory Balance objective is
\begin{equation}
\label{eq:continuous_subtb_loss}
    \mathcal{L}_{\mathrm{SubTB}}(x,y)
    =
    \mathbb{E}_{\tau\sim q_\varphi(\cdot\mid s_0)}
    \left[
    \sum\nolimits_{0\le i<j\le T}
    \chi_{i:j}^{2}
    \right].
\end{equation}
Minimizing this loss encourages the latent sampler to assign larger density to reasoning trajectories whose prefixes and completions obtain higher answer rewards, while preserving multiple valid reasoning paths instead of collapsing to a single mode.

\noindent{\textbf{Entropy-Weighted Subtrajectory Balance.}} Treating all subtrajectories uniformly is suboptimal: low-entropy paths are locally concentrated, while high-entropy paths usually correspond to regions with richer spread of information where the sampler needs stronger supervision~\citep{ag,zhu2025flowrl,du2025latent}. We therefore introduce \emph{Entropy-Weighted Subtrajectory Balance} (EW-SubTB), which uses the relative entropy of latent subtrajectories to reweight residuals without changing the reward-proportional objective. This entropy-aware weighting enables effective allocation of supervision across subtrajectories.

Given $S$ sampled trajectories for the same input $x$, we define the entropy for the latent $\mathbf{z}^{(s)}_{t+1}$ as
    $h_t^{(s)}
    =
    \mathcal H
    [
        q_\varphi(\mathbf z^{(s)}_{t+1}\mid s_t^{(s)})
    ]$
where $\mathcal{H}(\cdot)$ denotes differential entropy. For a subtrajectory $s_i\to\cdots\to s_j$, the length-normalized entropy is given by
    $\bar h_{i:j}^{(s)}
    =
    \frac{1}{j-i}
    \sum\nolimits_{t=i}^{j-1} h_t^{(s)}$. Thus, the entropy-aware weight is
\begin{equation}
\label{eq:entropy_weight}
    \omega_{i:j}^{(s)}
    =
    \operatorname{sg}
    \big[
    {
        \exp(\bar h_{i:j}^{(s)})
    }/\big({
        {1}/{|\mathcal S_{i:j}|} \cdot 
        \sum\nolimits_{r\in\mathcal S_{i:j}}
        \exp(\bar h_{i:j}^{(r)})\big)
    }
    \big],
\end{equation}
where $\mathcal S_{i:j}$ denotes the set of subtrajectories between indices $i$ and $j$ in $S$ sampled trajectories, and $\operatorname{sg}[\cdot]$ denotes the gradient stopping operation. This weighting keeps the average scale of the loss stable while assigning larger weights to relatively high-entropy latent reasoning subtrajectories.

For each sampled trajectory, the continuous SubTB residual is
\begin{equation}
\label{eq:advanced_subtb_residual}
{
\begin{aligned}
    \chi_{i:j}^{(s)}
    =
    \log
    \frac{
        \mathcal R_{x,y}(s_i^{(s)}\to\bot)
        \left[
        \prod_{t=i}^{j-1}
        q_\varphi(\mathbf z_{t+1}^{(s)}\mid s_t^{(s)})
        \bigl(1-\pi^\bot(s_t^{(s)})\bigr)
        \right]
        \pi^\bot(s_j^{(s)})
    }{
        \mathcal R_{x,y}(s_j^{(s)}\to\bot)
        \pi^\bot(s_i^{(s)})
    } .
\end{aligned}}
\end{equation}
The empirical EW-SubTB objective is
\begin{equation}
\label{eq:entropy_normalized_subtb}
    \mathcal L_{\mathrm{flow}}(x,y)
    =
    \frac{1}{S}
    \sum\nolimits_{s=1}^{S}
    \sum\nolimits_{0\le i<j\le T^{(s)}}
    \omega_{i:j}^{(s)}
    \left(
        \chi_{i:j}^{(s)}
    \right)^2.
\end{equation}
Since $\omega_{i:j}^{(s)}$ only reweights the squared residual and is not inserted into the balance ratio, EW-SubTB changes the optimization emphasis but preserves the same reward-proportional target distribution. Our proposed objective is validated via both empirical results (Sec.~\ref{sec:exp}) and analysis (Appendix~\ref{app:ex1}).

\subsection{Reference-Prior Regularization}
\label{sec:reference_prior}

Reward-proportional learning is difficult at the beginning of training because randomly explored continuous thoughts may not correspond to meaningful latent thoughts in the latent space. To ground early exploration, LTF uses a reference prior derived from teacher-generated reasoning traces when they are available~\citep{zhang2024multimodal,shao2024visualcot}. The prior anchors latent thoughts to meaningful regions of latent space, and the GFlowNet reward later reshapes the sampler toward concise and correct trajectories.

Given a reference latent rationale $r$ for input $x$, we introduce a reference branch after the decoder layer similar to the latent reasoning head. We define the reference transition density $p_{\theta'}^{\mathrm{ref}}$ as:
\begin{equation}
\label{eq:reference_prior_distribution}
p_{\theta'}^{\mathrm{ref}}(\mathbf{z}_t\mid s_{t-1})
    =
    \mathcal{N}\!\left(
        \mathbf{z}_t;
        \boldsymbol{\mu}_{\theta'}(s_{t-1}),
        \operatorname{diag}(\boldsymbol{\sigma}_{\theta'}^2(s_{t-1}))
    \right).
\end{equation}
The reference branch anchors the latent space semantically to optimize the LoRA attached to the backbone encoder layer. We regularize the prior sampler by aligning it to the reference prior:
\begin{equation}
\label{eq:prior_loss_revised}
    \mathcal{L}_{\mathrm{prior}}
    =
    - \mathbb{E}_{(x,r) \sim \mathcal{D}} \left[\log p_{\theta'}^{\mathrm{ref}}(r \mid  x) \right].
\end{equation}
Thus, early training benefits from semantically grounded latent states with stable exploration, while later training is governed by the accuracy--efficiency reward. If no rationale is available, this term is omitted and LTF trains only from answer supervision and the reward.

\subsection{Training Objective and Inference}
\label{sec:overall_training}

The training objective consists of complementary components for latent-thought sampling and answer prediction. The flow loss trains the sampler to allocate probability mass according to reward, while the cross entropy loss ensures that sampled thoughts remain effective for deriving the correct answer. Given an input-answer pair $(x,y)$, we define the answer loss as
\begin{equation}
	\label{eq:answer_loss_ltf}
	\mathcal{L}_{\mathrm{ans}}(x,y)
	=
	-
	\mathbb{E}_{\tau\sim q_\varphi(\cdot\mid H_x)}
	\left[
	\log p_\psi(y\mid \tau)
	\right].
\end{equation}
When computing the flow loss, we stop gradients through the scalar reward $\mathcal{R}_{x,y}(\tau)$; this prevents the decoder from changing the reward scale to satisfy balance and leaves answer learning to Eq.~\eqref{eq:answer_loss_ltf}.  The final objective is
\begin{equation}
	\label{eq:overall_loss_ltf}
	\mathcal{L}
	=
	\mathcal{L}_{\mathrm{flow}}
	+
	\lambda_{\mathrm{ans}}\mathcal{L}_{\mathrm{ans}}
	+
	\lambda_{\mathrm{prior}}\mathcal{L}_{\mathrm{prior}},
\end{equation}
where $\lambda_{\mathrm{ans}}$ and $\lambda_{\mathrm{prior}}$ are weighting hyperparameters. During inference, given the input $x$, LLM samples latent thoughts via Eq.~\eqref{eq:sample} and decodes the answer. By default, LTF uses a single trajectory for efficiency. With larger test-time budgets, it can adaptively sample multiple latent trajectories via tree of latent thoughts~\citep{yao2023tree} without decoding long explicit reasoning chains.

\section{Experiments}
\label{sec:exp}
\vspace{-0.2em}
\noindent{\textbf{Tasks, Metrics \& Baselines.}}
We evaluate LTF on multiple LLM backbones, including LLaMA-3.2 1B/3B~\citep{llama3}, LLaMA-3.1 8B~\citep{llama3}, and DeepSeek-R1-Distill-Qwen-1.5B~\citep{r1}. Following priors, we train and evaluate on GSM8K-Aug~\citep{deng2023implicit}, ASDiv-Aug~\citep{softcot}, and DU~\citep{du}, covering math reasoning, word problems, and data understanding. To test generalization and difficulty scalability, we further evaluate on out-of-domain math datasets, including GSM-Hard~\citep{gao2023pal}, SVAMP~\citep{patel2021nlp}, MultiArith~\citep{roy2015solving}, AQUA-RAT~\citep{ling2017program}, and MATH~\citep{hendrycks2021measuring}.  
We report two standard metrics~\citep{coconut,colar,regular}: Accuracy (Acc.), measuring the percentage of correct answers, and Reasoning Length (\#~L), measuring the number of reasoning steps per question. All results are averaged over five independent runs with different random seeds.  
We compare LTF with explicit reasoning methods, including CoT~\citep{cot} and Assist-CoT~\citep{softcot}, as well as latent reasoning methods, including iCoT~\citep{icot}, CODI~\citep{codi}, Coconut~\citep{coconut}, SoftCoT++~\citep{softcot2}, CoLaR~\citep{colar}, and ReGuLaR~\citep{regular}. All baselines follow their default configurations for fair comparison.

We train LTF with LoRA  using rank $r=128$ and scaling factor $\alpha=32$, and  AdamW~\citep{adamw} with learning rate $1\times10^{-4}$, weight decay $1\times10^{-2}$, global batch size 256, and 100 training epochs on RTX PRO 6000 GPUs. For fairness, all methods use the same training budget and stop once either the epoch limit or time budget is reached. We select the checkpoint with the best validation accuracy. During inference, generation uses temperature 0.9 and top-$p$ 0.95. See more details in Appendix~\ref{app:imp}.
\vspace{-0.3em}
\subsection{Main Results}
\label{sec:mainexp}
\vspace{-0.1em}
\begin{table*}[t!]
\centering
\caption{Performance comparison under finetuning setting across different backbones. We report the averaged Accuracy (Acc. \%) and Reasoning Length ($\#$ L).}
\label{tab:main}
\setlength{\tabcolsep}{10.5pt}
\scalebox{0.75}{
\begin{tabular}{cc||cccccc|cc}
\toprule
\multirow{2}{*}[-0.55ex]{\textbf{Model}} & \multirow{2}{*}[-0.55ex]{\textbf{Method}} & \multicolumn{2}{c}{\textbf{GSM8K-Aug}} & \multicolumn{2}{c}{\textbf{ASDiv-Aug}} & \multicolumn{2}{c}{\textbf{DU}} & \multicolumn{2}{c}{\textbf{Average}}\\
\cmidrule(lr){3-4} \cmidrule(lr){5-6} \cmidrule(lr){7-8} \cmidrule(lr){9-10}
& & \multicolumn{1}{c}{Acc.($\uparrow$)} & \multicolumn{1}{c}{\#~L($\downarrow$)} & \multicolumn{1}{c}{Acc.($\uparrow$)} & \multicolumn{1}{c}{\#~L($\downarrow$)} & \multicolumn{1}{c}{Acc.($\uparrow$)} & \multicolumn{1}{c}{\#~L($\downarrow$)} & \multicolumn{1}{c}{Acc.($\uparrow$)} & \multicolumn{1}{c}{\#~L($\downarrow$)} \\
\midrule
& CoT & 15.70 & 120.37 & 40.85 & 120.73 & 33.18 & 120.01 & 29.91 & 120.37 \\
& Assist-CoT & 16.30 & 119.85 & 42.18 & 119.53 & 35.34 & 119.58 & 31.27 & 119.65 \\
& SoftCoT++ & 17.04 & 119.73 & 45.72 & 119.48 & 38.18 & 119.70 & 33.65 & 119.64 \\
& iCoT & 19.80 & 0.00 & 46.36 & 0.00 & 41.37 & 0.00 & 35.84 & 0.00 \\
& CODI & 13.30 & 6.00 & 43.93 & 6.00 & 38.06 & 6.00 & 31.76 & 6.00 \\
& Coconut & 20.50 & 6.00 & 53.18 & 6.00 & 43.97 & 6.00 & 39.22 & 6.00 \\
& CoLaR & 26.60 & 5.63 & 67.19 & 5.19 & 50.12 & 4.17 & 47.97 & 5.00 \\
& ReGuLaR & 34.58 & 3.69 & 72.19 & 1.24 & 59.92 & 1.20 & 55.56 & 2.04 \\
\rowcolor{highlightred}\cellcolor{white}
\multirow{-8}{*}{\rotatebox{90}{\shortstack{\textbf{LLaMA-3.2}\\\textbf{Instruct 1B}}}} 
& LTF & \textbf{37.09} & \textbf{3.34} & \textbf{75.11} & \textbf{1.22} & \textbf{66.83} & \textbf{1.18} & \textbf{59.68} & \textbf{1.91} \\
\midrule
& CoT & 17.10 & 121.39 & 51.08 & 121.67 & 48.90 & 121.75 & 39.03	& 121.60 \\
& Assist-CoT & 17.90 & 120.98 & 52.42 & 120.11 & 50.28 & 120.04 & 40.20 & 120.38 \\
& SoftCoT++ & 18.59 & 120.77 & 56.39 & 119.97 & 53.31 & 119.85 & 42.76 & 120.20 \\
& iCoT & 21.57 & 0.00 & 58.45 & 0.00 & 54.18 & 0.00&44.73 & 0.00 \\
& Coconut & 32.71 & 6.00 & 66.76 & 6.00 & 57.23 & 6.00 & 52.23 & 6.00 \\
& CoLaR & 42.58 & 5.48 & 73.05 & 5.17 & 61.29 & 4.15 & 58.97 & 4.93 \\
& ReGuLaR & 45.59 & 3.89 & 79.46 & 1.24 & 70.31 & 1.18 & 65.12 & 2.10 \\
\rowcolor{highlightred}\cellcolor{white}
\multirow{-8}{*}{\rotatebox{90}{\shortstack{\textbf{LLaMA-3.2}\\\textbf{Instruct 3B}}}} 
& LTF & \textbf{48.30} & \textbf{3.42} & \textbf{84.04} & \textbf{1.21} & \textbf{76.82} & \textbf{1.17} & \textbf{68.85} & \textbf{1.95}\\ 
\midrule
& CoT & 19.17 & 121.30 & 89.23 & 121.93 & 65.67 & 121.04 & 58.02 & 121.42 \\
& Assist-CoT & 19.93 & 121.04 & 89.48 & 121.51 & 65.82 & 120.39 & 58.41 & 120.98 \\
& SoftCoT++ & 20.86 & 120.76 & 90.09 & 120.52 & 68.89 & 120.15 & 59.95 & 120.48 \\
& iCoT & 24.24 & 0.00 & 91.03 & 0.00 & 71.27 & 0.00 & 62.18 & 0.00 \\
& Coconut & 36.10 & 6.00 & 91.09 & 6.00 & 73.74 & 6.00 & 66.98 & 6.00 \\
& CoLaR & 45.19 & 5.62 & 91.12 & 5.51 & 76.59 & 4.18 & 70.97 & 5.10 \\
& ReGuLaR & 50.14 & 3.93 & 92.00 & 1.28 & 81.06 & \textbf{1.19} & 74.40 & 2.13 \\
\rowcolor{highlightred}\cellcolor{white}
\multirow{-8}{*}{\rotatebox{90}{\shortstack{\textbf{LLaMA-3.1}\\\textbf{Instruct 8B}}}} 
& LTF & \textbf{53.14} & \textbf{3.37} & \textbf{94.02} & \textbf{1.24} & \textbf{84.09} & 1.20 & \textbf{77.08} & \textbf{1.94} \\
\midrule
& CoT & 15.94 & 121.57 & 45.70 & 120.70 & 37.02 & 120.09 & 32.89 & 120.79 \\
& Assist-CoT & 17.36 & 120.81 & 46.48 & 119.81 & 38.62 & 119.67 & 34.15 & 120.10 \\
& SoftCoT++ & 18.19 & 120.64 & 48.51 & 119.93 & 43.04 & 119.83 & 36.58 & 120.13 \\
& iCoT & 18.87 & 0.00 & 48.53 & 0.00 & 43.18 & 0.00 & 36.86 & 0.00 \\
& Coconut & 24.91 & 6.00 & 56.85 & 6.00 & 46.46 & 6.00 & 42.74 & 6.00 \\
& CoLaR & 27.38 & 5.71 & 69.36 & 5.23 & 50.35 & 4.17 & 49.03 & 5.04 \\
& ReGuLaR & 34.69 & 3.54 & 75.59 & 1.21 & 63.91 & 1.19 & 58.06 & 1.98 \\
\rowcolor{highlightred}\cellcolor{white}
\multirow{-8}{*}{\rotatebox{90}{\shortstack{\textbf{DeepSeek-R1}\\\textbf{Distill-Qwen 1.5B}}}} 
& LTF & \textbf{36.94} & \textbf{3.27} & \textbf{79.75} & \textbf{1.20} & \textbf{69.80} & \textbf{1.17} & \textbf{62.16} & \textbf{1.88} \\
\bottomrule
\end{tabular}}
\vspace{-0.5em}
\end{table*}
\noindent{\textbf{Results of Finetuning.}}
Table~\ref{tab:main} compares LTF with explicit and latent reasoning baselines on three mathematical reasoning datasets. LTF achieves the best accuracy--efficiency tradeoffs across tasks. Compared with the strongest baseline ReGuLaR, LTF (LLaMA-1B) improves the average accuracy from $34.58\%$ to $37.09\%$, while reducing the average reasoning length from $3.69$ to $3.34$ on the challenging GSM8K-Aug task. This suggests that LTF is effective for problems requiring stronger multi-step arithmetic reasoning, where compact latent transitions can preserve task-relevant information. In addition, LTF (LLaMA-8B) improves the average accuracy from $50.14\%$ to $53.14\%$, while reducing the average reasoning length from $3.93$ to $3.37$, showing the robust scalability across backbones. On the other tasks, LTF adaptively allocates reasoning budgets for accuracy gains. On ASDiv-Aug, LTF improves the average accuracy from $92.00\%$ to $94.02\%$, reducing reasoning length from $1.28$ to $1.24$. We attribute the performance to the entropy-weighted SubTB objective, which encourages compact and informative latent thought states rather than redundant reasoning trajectories.

\begin{table*}[t!]
\centering
\caption{Extreme compression performance of LTF under finetuning setting. We report the averaged Accuracy (Acc. \%) and Reasoning Length ($\#$ L).}
\label{tab:extreme-compression}
\setlength{\tabcolsep}{6pt}
\scalebox{0.75}{
\begin{tabular}{clcccccccc|cc}
\toprule
\multirow{2}{*}[-0.55ex]{\textbf{Dataset}} &\multirow{2}{*}[-0.55ex]{\textbf{Method}} & \multicolumn{2}{c}{\textbf{LLaMA-1B}} & \multicolumn{2}{c}{\textbf{LLaMA-3B}} & \multicolumn{2}{c}{\textbf{LLaMA-8B}} & \multicolumn{2}{c}{\textbf{DS-1.5B}} & \multicolumn{2}{c}{\textbf{Average}} \\
\cmidrule(lr){3-4} \cmidrule(lr){5-6} \cmidrule(lr){7-8} \cmidrule(lr){9-10} \cmidrule(lr){11-12} 
& & \multicolumn{1}{c}{Acc.($\uparrow$)} & \multicolumn{1}{c}{\#~L($\downarrow$)} & \multicolumn{1}{c}{Acc.($\uparrow$)} & \multicolumn{1}{c}{\#~L($\downarrow$)} & \multicolumn{1}{c}{Acc.($\uparrow$)} & \multicolumn{1}{c}{\#~L($\downarrow$)} & \multicolumn{1}{c}{Acc.($\uparrow$)} & \multicolumn{1}{c}{\#~L($\downarrow$)}  & \multicolumn{1}{c}{Acc.($\uparrow$)} & \multicolumn{1}{c}{\#~L($\downarrow$)} \\
\midrule
& Coconut & 5.18 & 6.00 & 7.72 & 6.00 & 8.68 & 6.00 & 6.67 & 6.00 & 7.06 & 6.00 \\
& CoLaR & 5.31 & 58.29 & 8.35 & 60.44 & 9.13 & 67.19 & 11.13 & 62.20 & 8.48 & 62.03 \\
& ReGuLaR & 6.62 & 1.00 & 11.80 & 1.00 & 13.90 & 1.00 & 15.60 & 1.00 & 11.98 & 1.00\\
\rowcolor{highlightred}
\cellcolor{white}\multirow{-3}{*}{MATH}
& LTF & $\textbf{8.10}$ & $\textbf{1.00}$ & $\textbf{16.18}$ & $\textbf{1.00}$ & $\textbf{17.31}$ & $\textbf{1.00}$ & $\textbf{17.20}$ & $\textbf{1.00}$ & \textbf{14.70} & \textbf{1.00} \\ 
\midrule
& Coconut & 23.17& 6.00 & 26.31 & 6.00 & 30.94 & 6.00 & 27.81 & 6.00 & 27.06& 6.00 \\
& CoLaR & 23.69 & 18.60 & 32.78 & 28.20 & 34.52 & 20.30 & 34.33 & 27.00 & 31.33 & 23.53 \\
& ReGuLaR & 37.28 & 1.00 & 39.12 & 1.00 & 42.41 & 1.00 & 39.07 & 1.00 & 39.47 & 1.00\\
\rowcolor{highlightred}
\cellcolor{white}\multirow{-3}{*}{AQUA-RAT}
& LTF & $\textbf{39.43}$ & $\textbf{1.00}$ & $\textbf{43.18}$ & $\textbf{1.00}$ & $\textbf{46.62}$ & $\textbf{1.00}$ & $\textbf{43.07}$ & $\textbf{1.00}$ & \textbf{43.08} & \textbf{1.00} \\ 
\bottomrule
\end{tabular}
}
\end{table*}
\noindent{\textbf{Extreme Compression Evaluation.}}
Although accuracy decreases under extreme compression, Table~\ref{tab:extreme-compression} shows that LTF consistently outperforms baselines across all settings on both LLM backbones, verifying its advantage in preserving semantic information. Compared with ReGuLaR, LTF improves the average accuracy by $2.72\%$ on MATH and $3.61\%$ on AQUA-RAT, demonstrating its superior ability to preserve task-relevant semantic information under compression.

\begin{table*}[t!]
\centering
\caption{Performance comparison under transfer learning setting  across different backbones. We report the averaged Accuracy (Acc. \%) and Reasoning Length ($\#$ L).}
\label{tab:transfer}
\setlength{\tabcolsep}{10.5pt}
\scalebox{0.75}{
\begin{tabular}{cc||cccccc|cc}
\toprule
\multirow{2}{*}[-0.55ex]{\textbf{Model}} & \multirow{2}{*}[-0.55ex]{\textbf{Method}} & \multicolumn{2}{c}{\textbf{GSM-Hard}} & \multicolumn{2}{c}{\textbf{SVAMP}} & \multicolumn{2}{c}{\textbf{MultiArith}} & \multicolumn{2}{c}{\textbf{Average}}\\
\cmidrule(lr){3-4} \cmidrule(lr){5-6} \cmidrule(lr){7-8} \cmidrule(lr){9-10}
& & \multicolumn{1}{c}{Acc.($\uparrow$)} & \multicolumn{1}{c}{\#~L($\downarrow$)} & \multicolumn{1}{c}{Acc.($\uparrow$)} & \multicolumn{1}{c}{\#~L($\downarrow$)} & \multicolumn{1}{c}{Acc.($\uparrow$)} & \multicolumn{1}{c}{\#~L($\downarrow$)} & \multicolumn{1}{c}{Acc.($\uparrow$)} & \multicolumn{1}{c}{\#~L($\downarrow$)} \\
\midrule
& CoT & 2.74 & 120.46 & 27.17 & 120.80 & 28.29 & 120.14 & 19.40 & 120.47 \\
& Assist-CoT & 3.02 & 125.19 & 28.04 & 123.02 & 30.06 & 121.86 & 20.37 & 123.36 \\
& SoftCoT++ & 3.77 & 123.56 & 30.96 & 123.42 & 32.13 & 120.66 & 22.29 & 122.55 \\
& iCoT & 3.82 & 0.00 & 36.42 & 0.00 & 38.19 & 0.00 & 26.14 & 0.00 \\
& CODI & 2.94 & 6.00 & 21.65 & 6.00 & 19.20 & 6.00 & 14.60 & 6.00 \\
& Coconut & 4.87 & 6.00 & 38.90 & 6.00 & 41.30 & 6.00 & 28.36 & 6.00 \\
& CoLaR & 6.22 & 7.03 & 47.22 & 2.94 & 86.93 & 3.22 & 46.79 & 4.40 \\
& ReGuLaR & 8.25 & 4.10 & 48.19 & 2.10 & 88.97 & 2.29 & 48.47 & 2.83 \\
\rowcolor{highlightred}\cellcolor{white}
\multirow{-8}{*}{\rotatebox{90}{\shortstack{\textbf{LLaMA-3.2}\\\textbf{Instruct 1B}}}} 
& LTF & \textbf{8.71} & \textbf{3.97} & \textbf{52.01} & \textbf{2.11} & \textbf{90.37} & \textbf{2.17} & \textbf{50.36} & \textbf{2.75} \\
\midrule
& CoT & 3.84 & 121.68 & 34.01 & 121.52 & 35.16 & 121.54 & 24.34 & 121.58 \\
& Assist-CoT & 4.16 & 125.03 & 35.06 & 123.42 & 37.03 & 121.98 & 25.42 & 123.48 \\
& SoftCoT++ & 5.21 & 124.72 & 38.89 & 122.96 & 40.26 & 121.74 & 28.12 & 123.14 \\
& iCoT & 5.19 & 0.00 & 44.98 & 0.00 & 46.42 & 0.00 & 32.19 & 0.00 \\
& Coconut & 7.34 & 6.00 & 52.47 & 6.00 & 58.79 & 6.00 & 39.53 & 6.00 \\
& CoLaR & 9.73 & 6.46 & 60.70 & 2.72 & 90.10 & 3.04 & 53.51 & 4.07 \\
& ReGuLaR & 11.20 & 3.73 & 62.32 & 2.23 & 91.27 & 2.17 & 54.93 & 2.71 \\
\rowcolor{highlightred}\cellcolor{white}
\multirow{-8}{*}{\rotatebox{90}{\shortstack{\textbf{LLaMA-3.2}\\\textbf{Instruct 3B}}}} 
& LTF & \textbf{11.88} & \textbf{3.64} & \textbf{65.48} & \textbf{2.14} & \textbf{92.94} & \textbf{2.10} & \textbf{56.77} & \textbf{2.63} \\
\midrule
& CoT & 4.51 & 121.66 & 37.43 & 121.56 & 38.82 & 121.84 & 26.92 & 121.69 \\
& Assist-CoT & 4.75 & 125.59 & 38.62 & 124.62 & 40.60 & 122.64 & 27.99 & 124.28 \\
& SoftCoT++ & 5.91 & 124.82 & 42.85 & 123.82 & 43.19 & 121.98 & 30.65 & 123.54 \\
& iCoT & 5.82 & 0.00 & 49.48 & 0.00 & 50.39 & 0.00 & 35.23 & 0.00 \\
& Coconut & 8.26 & 6.00 & 57.18 & 6.00 & 62.41 & 6.00 & 42.62 & 6.00 \\
& CoLaR & 10.97 & 6.53 & 61.30 & 2.83 & 93.80 & 3.14 & 55.36 & 4.17 \\
& ReGuLaR & 12.53 & 3.92 & 68.54 & \textbf{2.07} & 95.28 & 2.26 & 58.78 & 2.75 \\
\rowcolor{highlightred}\cellcolor{white}
\multirow{-8}{*}{\rotatebox{90}{\shortstack{\textbf{LLaMA-3.1}\\\textbf{Instruct 8B}}}} 
& LTF & \textbf{13.19} & \textbf{3.70} & \textbf{72.29} & 2.09 & \textbf{97.18} & \textbf{2.16} & \textbf{60.89} & \textbf{2.65} \\ 
\midrule
& CoT & 2.81 & 120.78 & 27.23 & 121.48 & 28.33 & 121.53 & 19.46 & 121.26 \\
& Assist-CoT & 3.07 & 126.03 & 28.12 & 123.76 & 30.14 & 122.70 & 20.44 & 124.16 \\
& SoftCoT++ & 3.86 & 123.72 & 31.02 & 123.26 & 32.18 & 121.92 & 22.35 & 122.97 \\
& iCoT & 3.83 & 0.00 & 36.49 & 0.00 & 38.24 & 0.00 & 26.19 & 0.00 \\
& Coconut & 4.96 & 6.00 & 38.98 & 6.00 & 41.31 & 6.00 & 28.41 & 6.00 \\
& CoLaR & 6.32 & 7.03 & 47.27 & 2.94 & 87.01 & 3.22 & 46.86 & 4.40 \\
& ReGuLaR & 8.33 & 4.10 & 48.20 & 2.10 & 89.04 & 2.29 & 48.52 & 2.83 \\
\rowcolor{highlightred}\cellcolor{white}
\multirow{-8}{*}{\rotatebox{90}{\shortstack{\textbf{DeepSeek-R1}\\\textbf{Distill-Qwen 1.5B}}}} 
& LTF & \textbf{8.83} & \textbf{3.94} & \textbf{52.76} & \textbf{2.08} & \textbf{91.10} & \textbf{2.21} & \textbf{50.90} & \textbf{2.74} \\
\bottomrule
\end{tabular}}
\vspace{-1em}
\end{table*}
\noindent{\textbf{Results of Transfer Learning.}}
Table~\ref{tab:transfer} compares LTF with baseline methods on transfer learning tasks across three datasets and four backbones. LTF achieves the highest average accuracy for all backbones, improving over ReGuLaR by $1.89\%$, $1.84\%$, $2.11\%$, and $2.38\%$ on LLaMA-1B, LLaMA-3B, LLaMA-8B, and DS-1.5B, respectively. Meanwhile, LTF maintains a more concise reasoning length than ReGuLaR and explicit CoT in all settings. These results validate the reasoning generalization ability of LTF, improving broader reasoning performance while preserving strong compression efficiency. Additional results with 95\% confidence interval are provided in Appendix~\ref{app:ex}.

\vspace{-0.4em}
\subsection{Ablation Studies}
\vspace{-0.3em}
\label{sec:ablation}
\begin{table*}[t!]
\centering
\caption{Ablation on entropy weighting and training rollout sample size $S$ in LTF using the LLaMA-3.1 Instruct 1B backbone. We report the averaged Accuracy (Acc. \%) and Reasoning Length ($\#$ L).}
\label{tab:ablation}
\setlength{\tabcolsep}{8.5pt}
\scalebox{0.8}{
\begin{tabular}{cc|cccccccc}
\toprule
\multicolumn{2}{c|}{\textbf{Settings}} & \multicolumn{2}{c}{\textbf{GSM8K-Aug}} & \multicolumn{2}{c}{\textbf{ASDiv-Aug}} & \multicolumn{2}{c}{\textbf{DU}} & \multicolumn{2}{c}{\textbf{Average}}\\
\cmidrule(lr){1-2} \cmidrule(lr){3-4} \cmidrule(lr){5-6} \cmidrule(lr){7-8} \cmidrule(lr){9-10}
Entropy Weighting & $S$ & Acc.($\uparrow$) & \#~L($\downarrow$) & Acc.($\uparrow$) & \#~L($\downarrow$) & Acc.($\uparrow$) & \#~L($\downarrow$) & Acc.($\uparrow$) & \#~L($\downarrow$) \\
\midrule
- & 5 & 34.78 & 3.31 & 72.74 & 1.20 & 63.50 & 1.19 & 57.01 & 1.90 \\
\ding{51} & 5 & 35.19 & 3.33 & 73.42 & 1.23 & 63.61 & 1.19 & 57.41 & 1.92 \\
\midrule
- & 10 & 36.10 & 3.30 & 74.08 & 1.19 & 65.02 & 1.19 & 58.40 & 1.89 \\
\ding{51} & 10 & 36.71 & 3.33 & 74.65 & 1.21 & 65.84 & 1.18 & 59.07 & 1.91 \\
\midrule
- & 20 & 36.64	&3.31&	74.80	&1.20	&64.74&	1.19	&58.73&	1.90\\
\ding{51} & 20 & 37.09 & 3.34 & 75.11 & 1.22 & 66.83 & 1.18 & 59.68 & 1.91 \\
\bottomrule
\end{tabular}}
\vspace{-0.1em}
\end{table*}

We conduct ablation studies on the key components of LTF. Due to the space limitation, we defer the analyses of reference-prior regularization and test-time scaling to Appendix~\ref{app:ex}.

\noindent{\textbf{Analysis on Hyperparameters.}}
Table~\ref{tab:ablation} studies entropy weighting and rollout sample size $S$ in LTF. \textbf{1)} Entropy weighting consistently improves accuracy with little change in reasoning length, and its gains increase with larger $S$ ($+0.40\%$, $+0.67\%$, and $+0.95\%$ for $S=5,10,20$), suggesting stronger benefits when sampled subtrajectories are more diverse. It enables effective credit assignment to improve reasoning performance over uniform SubTB. \textbf{2)} Increasing $S$ improves performance by expanding trajectory diversity: with entropy weighting, average accuracy rises from $57.41\%$ to $59.68\%$ as $S$ increases from $5$ to $20$, showing the scaling effect during training.

\begin{table*}[t!]
\centering
\caption{Ablation on latent thought exploration objective using the LLaMA-3.1 Instruct 1B backbone. We report the averaged Accuracy (Acc. \%) and Reasoning Length ($\#$ L).}
\label{tab:ablation1}
\setlength{\tabcolsep}{12.5pt}
\scalebox{0.8}{
\begin{tabular}{c|cccccccc}
\toprule
\multirow{2}{*}[-0.55ex]{\textbf{Model}} & \multicolumn{2}{c}{\textbf{GSM8K-Aug}} & \multicolumn{2}{c}{\textbf{ASDiv-Aug}} & \multicolumn{2}{c}{\textbf{DU}} & \multicolumn{2}{c}{\textbf{Average}}\\
\cmidrule(lr){2-3} \cmidrule(lr){4-5} \cmidrule(lr){6-7} \cmidrule(lr){8-9}
& Acc.($\uparrow$) & \#~L($\downarrow$) & Acc.($\uparrow$) & \#~L($\downarrow$) & Acc.($\uparrow$) & \#~L($\downarrow$) & Acc.($\uparrow$) & \#~L($\downarrow$) \\
\midrule
GRPO & 25.93 & 13.17 & 66.93 & 12.45 & 49.62 & 11.14 & 47.49 & 12.25\\
DB   & 35.17 & 7.73  & 72.28 & 7.19  & 60.48 & 6.92  & 55.98 & 7.28\\
TB   & 35.84 & 8.01  & 73.05 & 7.39  & 61.52 & 7.14  & 56.80 & 7.51\\
\midrule
LTF & 37.09 & 3.34 & 75.11 & 1.22 & 66.83 & 1.18 & 59.68 & 1.91 \\
\bottomrule
\end{tabular}}
\vspace{-1em}
\end{table*}

\vspace{-0.3em}
\noindent{\textbf{Analysis on Latent Thought Exploration Objective.}}
Table~\ref{tab:ablation1} studies the latent thought exploration objective in LTF. Baseline details are provided in Appendix~\ref{app:baseline}. \textbf{1)} Compared with GRPO, GFlowNet-based objectives achieve substantially better accuracy--length tradeoffs. DB improves the average accuracy from $47.49\%$ to $55.98\%$ while reducing the average reasoning length from $12.25$ to $7.28$, suggesting that flow-based learning provides more structured supervision for latent thought exploration than standard RL. \textbf{2)} LTF achieves the best overall tradeoff. It obtains the highest average accuracy of $59.68\%$ while reducing the average reasoning length to only $1.91$. These results show that LTF better aligns latent thought exploration with both reasoning effectiveness and conciseness.

\vspace{-0.5em}
\section{Conclusion}
\vspace{-0.5em}
We introduce LTF, a framework for efficient latent reasoning in LLMs. LTF models reasoning as variable-length continuous latent trajectories and trains the sampler with a continuous GFlowNet objective, allocating probability mass according to answer quality and computational cost. An entropy-weighted subtrajectory balance objective and a reference-prior regularizer further improve exploration and stability. Extensive experiments highlight the superiority of LTF over explicit and latent reasoning methods, showing a path towards a better reasoning accuracy--efficiency frontier.
\vspace{-0.3em}

\noindent{\textbf{Limitations.}}
Our experiments primarily focus on textual tasks involving words and symbols; extending to other modalities, such as vision and speech, remains as future work. In addition, we will further explore the generalization ability of Latent Thought Flow in theory.

\bibliography{reference}

@article{lu2022learn,
  title={Learn to explain: Multimodal reasoning via thought chains for science question answering},
  author={Lu, Pan and Mishra, Swaroop and Xia, Tanglin and Qiu, Liang and Chang, Kai-Wei and Zhu, Song-Chun and Tafjord, Oyvind and Clark, Peter and Kalyan, Ashwin},
  journal={Advances in neural information processing systems},
  volume={35},
  pages={2507--2521},
  year={2022}
}

@article{yang2023mmreact,
  title={Mm-react: Prompting chatgpt for multimodal reasoning and action},
  author={Yang, Zhengyuan and Li, Linjie and Wang, Jianfeng and Lin, Kevin and Azarnasab, Ehsan and Ahmed, Faisal and Liu, Zicheng and Liu, Ce and Zeng, Michael and Wang, Lijuan},
  journal={arXiv preprint arXiv:2303.11381},
  year={2023}
}

@inproceedings{suris2023viper,
  title={Vipergpt: Visual inference via python execution for reasoning},
  author={Sur{\'\i}s, D{\'\i}dac and Menon, Sachit and Vondrick, Carl},
  booktitle={Proceedings of the IEEE/CVF international conference on computer vision},
  pages={11888--11898},
  year={2023}
}

@article{kong2025latent,
  title={Latent thought models with variational bayes inference-time computation},
  author={Kong, Deqian and Zhao, Minglu and Xu, Dehong and Pang, Bo and Wang, Shu and Honig, Edouardo and Si, Zhangzhang and Li, Chuan and Xie, Jianwen and Xie, Sirui and others},
  journal={arXiv preprint arXiv:2502.01567},
  year={2025}
}

@article{kojima2022large,
  title={Large language models are zero-shot reasoners},
  author={Kojima, Takeshi and Gu, Shixiang Shane and Reid, Machel and Matsuo, Yutaka and Iwasawa, Yusuke},
  journal={Advances in neural information processing systems},
  volume={35},
  pages={22199--22213},
  year={2022}
}

@article{wang2023selfconsistency,
  title={Self-consistency improves chain of thought reasoning in language models},
  author={Wang, Xuezhi and Wei, Jason and Schuurmans, Dale and Le, Quoc and Chi, Ed and Narang, Sharan and Chowdhery, Aakanksha and Zhou, Denny},
  journal={arXiv preprint arXiv:2203.11171},
  year={2022}
}

@article{yao2023tree,
  title={Tree of thoughts: Deliberate problem solving with large language models},
  author={Yao, Shunyu and Yu, Dian and Zhao, Jeffrey and Shafran, Izhak and Griffiths, Tom and Cao, Yuan and Narasimhan, Karthik},
  journal={Advances in neural information processing systems},
  volume={36},
  pages={11809--11822},
  year={2023}
}

@article{zhang2024multimodal,
  title={Multimodal chain-of-thought reasoning in language models},
  author={Zhang, Zhuosheng and Zhang, Aston and Li, Mu and Zhao, Hai and Karypis, George and Smola, Alex},
  journal={arXiv preprint arXiv:2302.00923},
  year={2023}
}

@article{shao2024visualcot,
  title={Visual cot: Advancing multi-modal language models with a comprehensive dataset and benchmark for chain-of-thought reasoning},
  author={Shao, Hao and Qian, Shengju and Xiao, Han and Song, Guanglu and Zong, Zhuofan and Wang, Letian and Liu, Yu and Li, Hongsheng},
  journal={Advances in Neural Information Processing Systems},
  volume={37},
  pages={8612--8642},
  year={2024}
}

@article{goyal2024pause,
  title={Think before you speak: Training language models with pause tokens},
  author={Goyal, Sachin and Ji, Ziwei and Rawat, Ankit Singh and Menon, Aditya Krishna and Kumar, Sanjiv and Nagarajan, Vaishnavh},
  journal={arXiv preprint arXiv:2310.02226},
  year={2023}
}

@article{cheng2024compressed,
  title={Compressed chain of thought: Efficient reasoning through dense representations},
  author={Cheng, Jeffrey and Van Durme, Benjamin},
  journal={arXiv preprint arXiv:2412.13171},
  year={2024}
}

@inproceedings{ramji2026abstract,
  title={Learning Efficient Latent Reasoning with Abstract Chain-of-Thought},
  author={Ramji, Keshav and Naseem, Tahira and Astudillo, Ram{\'o}n Fernandez},
  booktitle={Workshop on Latent $\{$$\backslash$\&$\}$ Implicit Thinking $\{$$\backslash$textendash$\}$ Going Beyond CoT Reasoning},
  year={2026}
}

@article{hu2024amortizing,
  title={Amortizing intractable inference in large language models},
  author={Hu, Edward J and Jain, Moksh and Elmoznino, Eric and Kaddar, Younesse and Lajoie, Guillaume and Bengio, Yoshua and Malkin, Nikolay},
  journal={arXiv preprint arXiv:2310.04363},
  year={2023}
}

@article{takase2024gflownet,
  title={Gflownet fine-tuning for diverse correct solutions in mathematical reasoning tasks},
  author={Takase, Ryoichi and Tsunokake, Masaya and Tsuchiya, Yuta and Inuzuka, Shota},
  journal={arXiv preprint arXiv:2410.20147},
  year={2024}
}

@inproceedings{kang2025gflowvlm,
  title={Gflowvlm: Enhancing multi-step reasoning in vision-language models with generative flow networks},
  author={Kang, Haoqiang and Sachdeva, Enna and Gupta, Piyush and Bae, Sangjae and Lee, Kwonjoon},
  booktitle={Proceedings of the Computer Vision and Pattern Recognition Conference},
  pages={3815--3825},
  year={2025}
}

@article{kingma2014autoencoding,
  title={Auto-encoding variational bayes},
  author={Kingma, Diederik P and Welling, Max},
  journal={arXiv preprint arXiv:1312.6114},
  year={2013}
}

@article{colar,
  title={Think silently, think fast: Dynamic latent compression of llm reasoning chains},
  author={Tan, Wenhui and Li, Jiaze and Ju, Jianzhong and Luo, Zhenbo and Song, Ruihua and Luan, Jian},
  journal={arXiv preprint arXiv:2505.16552},
  year={2025}
}

@article{regular,
  title={ReGuLaR: Variational Latent Reasoning Guided by Rendered Chain-of-Thought},
  author={Wang, Fanmeng and Liu, Haotian and Zhao, Guojiang and Xu, Hongteng and Gao, Zhifeng},
  journal={arXiv preprint arXiv:2601.23184},
  year={2026}
}

@article{gflownet,
  title={Flow network based generative models for non-iterative diverse candidate generation},
  author={Bengio, Emmanuel and Jain, Moksh and Korablyov, Maksym and Precup, Doina and Bengio, Yoshua},
  journal={Advances in neural information processing systems},
  volume={34},
  pages={27381--27394},
  year={2021}
}

@inproceedings{cgflow,
  title={A theory of continuous generative flow networks},
  author={Lahlou, Salem and Deleu, Tristan and Lemos, Pablo and Zhang, Dinghuai and Volokhova, Alexandra and Hern{\'a}ndez-Garc{\i}a, Alex and Ezzine, L{\'e}na N{\'e}hale and Bengio, Yoshua and Malkin, Nikolay},
  booktitle={International Conference on Machine Learning},
  pages={18269--18300},
  year={2023},
  organization={PMLR}
}

@article{gthink,
  title={Flow of reasoning: Training llms for divergent reasoning with minimal examples},
  author={Yu, Fangxu and Jiang, Lai and Kang, Haoqiang and Hao, Shibo and Qin, Lianhui},
  journal={arXiv preprint arXiv:2406.05673},
  year={2024}
}

@article{gthink1,
  title={Latent chain-of-thought for visual reasoning},
  author={Sun, Guohao and Hua, Hang and Wang, Jian and Luo, Jiebo and Dianat, Sohail and Rabbani, Majid and Rao, Raghuveer and Tao, Zhiqiang},
  journal={arXiv preprint arXiv:2510.23925},
  year={2025}
}

@inproceedings{ag,
  title={When do GFlowNets learn the right distribution?},
  author={Silva, Tiago and Alves, Rodrigo Barreto and da Silva, Eliezer de Souza and Souza, Amauri H and Garg, Vikas and Kaski, Samuel and Mesquita, Diego},
  booktitle={The Thirteenth International Conference on Learning Representations},
  year={2025}
}

@inproceedings{subtb,
  title={Learning gflownets from partial episodes for improved convergence and stability},
  author={Madan, Kanika and Rector-Brooks, Jarrid and Korablyov, Maksym and Bengio, Emmanuel and Jain, Moksh and Nica, Andrei Cristian and Bosc, Tom and Bengio, Yoshua and Malkin, Nikolay},
  booktitle={International Conference on Machine Learning},
  pages={23467--23483},
  year={2023},
  organization={PMLR}
}

@article{tb,
  title={Trajectory balance: Improved credit assignment in gflownets},
  author={Malkin, Nikolay and Jain, Moksh and Bengio, Emmanuel and Sun, Chen and Bengio, Yoshua},
  journal={Advances in Neural Information Processing Systems},
  volume={35},
  pages={5955--5967},
  year={2022}
}

@article{gflownet1,
  title={Gflownet foundations},
  author={Bengio, Yoshua and Lahlou, Salem and Deleu, Tristan and Hu, Edward J and Tiwari, Mo and Bengio, Emmanuel},
  journal={Journal of Machine Learning Research},
  volume={24},
  number={210},
  pages={1--55},
  year={2023}
}

@article{coconut,
  title={Training large language models to reason in a continuous latent space},
  author={Hao, Shibo and Sukhbaatar, Sainbayar and Su, DiJia and Li, Xian and Hu, Zhiting and Weston, Jason and Tian, Yuandong},
  journal={arXiv preprint arXiv:2412.06769},
  year={2024}
}

@inproceedings{softcot,
  title={Softcot: Soft chain-of-thought for efficient reasoning with llms},
  author={Xu, Yige and Guo, Xu and Zeng, Zhiwei and Miao, Chunyan},
  booktitle={Proceedings of the 63rd Annual Meeting of the Association for Computational Linguistics (Volume 1: Long Papers)},
  pages={23336--23351},
  year={2025}
}

@article{softcot2,
  title={Softcot++: Test-time scaling with soft chain-of-thought reasoning},
  author={Xu, Yige and Guo, Xu and Zeng, Zhiwei and Miao, Chunyan},
  journal={arXiv preprint arXiv:2505.11484},
  year={2025}
}

@article{icot,
  title={From explicit cot to implicit cot: Learning to internalize cot step by step},
  author={Deng, Yuntian and Choi, Yejin and Shieber, Stuart},
  journal={arXiv preprint arXiv:2405.14838},
  year={2024}
}

@inproceedings{codi,
  title={Codi: Compressing chain-of-thought into continuous space via self-distillation},
  author={Shen, Zhenyi and Yan, Hanqi and Zhang, Linhai and Hu, Zhanghao and Du, Yali and He, Yulan},
  booktitle={Proceedings of the 2025 Conference on Empirical Methods in Natural Language Processing},
  pages={677--693},
  year={2025}
}

@article{cot,
  title={Chain-of-thought prompting elicits reasoning in large language models},
  author={Wei, Jason and Wang, Xuezhi and Schuurmans, Dale and Bosma, Maarten and Xia, Fei and Chi, Ed and Le, Quoc V and Zhou, Denny and others},
  journal={Advances in neural information processing systems},
  volume={35},
  pages={24824--24837},
  year={2022}
}

@article{deng2023implicit,
  title={Implicit chain of thought reasoning via knowledge distillation},
  author={Deng, Yuntian and Prasad, Kiran and Fernandez, Roland and Smolensky, Paul and Chaudhary, Vishrav and Shieber, Stuart},
  journal={arXiv preprint arXiv:2311.01460},
  year={2023}
}

@inproceedings{gao2023pal,
  title={Pal: Program-aided language models},
  author={Gao, Luyu and Madaan, Aman and Zhou, Shuyan and Alon, Uri and Liu, Pengfei and Yang, Yiming and Callan, Jamie and Neubig, Graham},
  booktitle={International conference on machine learning},
  pages={10764--10799},
  year={2023},
  organization={PMLR}
}

@inproceedings{patel2021nlp,
  title={Are NLP models really able to solve simple math word problems?},
  author={Patel, Arkil and Bhattamishra, Satwik and Goyal, Navin},
  booktitle={Proceedings of the 2021 conference of the North American chapter of the association for computational linguistics: human language technologies},
  pages={2080--2094},
  year={2021}
}

@inproceedings{roy2015solving,
  title={Solving general arithmetic word problems},
  author={Roy, Subhro and Roth, Dan},
  booktitle={Proceedings of the 2015 conference on empirical methods in natural language processing},
  pages={1743--1752},
  year={2015}
}

@inproceedings{ling2017program,
  title={Program induction by rationale generation: Learning to solve and explain algebraic word problems},
  author={Ling, Wang and Yogatama, Dani and Dyer, Chris and Blunsom, Phil},
  booktitle={Proceedings of the 55th annual meeting of the association for computational linguistics (volume 1: Long papers)},
  pages={158--167},
  year={2017}
}

@article{hendrycks2021measuring,
  title={Measuring mathematical problem solving with the math dataset},
  author={Hendrycks, Dan and Burns, Collin and Kadavath, Saurav and Arora, Akul and Basart, Steven and Tang, Eric and Song, Dawn and Steinhardt, Jacob},
  journal={arXiv preprint arXiv:2103.03874},
  year={2021}
}

@article{du,
  title={Beyond the imitation game: Quantifying and extrapolating the capabilities of language models},
  author={Srivastava, Aarohi and Rastogi, Abhinav and Rao, Abhishek and Shoeb, Abu Awal Md and Abid, Abubakar and Fisch, Adam and Brown, Adam R and Santoro, Adam and Gupta, Aditya and Garriga-Alonso, Adri{\`a} and others},
  journal={Transactions on machine learning research},
  year={2023}
}

@inproceedings{jiangrethinking,
  title={Rethinking LLM Reasoning: From Explicit Trajectories to Latent Representations},
  author={Jiang, Cong and Zhang, Xiaofeng and Zhu, Fangzhi and Chen, XiaoWei and Zhu, Junxiong and Zhang, Zheng},
  booktitle={The Fourteenth International Conference on Learning Representations},
  year={2026}
}

@article{sun2026llm,
  title={LLM Reasoning as Trajectories: Step-Specific Representation Geometry and Correctness Signals},
  author={Sun, Lihao and Dong, Hang and Qiao, Bo and Lin, Qingwei and Zhang, Dongmei and Rajmohan, Saravan},
  journal={arXiv preprint arXiv:2604.05655},
  year={2026}
}

@article{zhou2026lepo,
  title={LEPO: Latent Reasoning Policy Optimization for Large Language Models},
  author={Zhou, Yuyan and Yu, Jiarui and Dong, Hande and Hao, Zhezheng and Wang, Hong and Zhang, Jianqing and Lin, Qiang},
  journal={arXiv e-prints},
  pages={arXiv--2604},
  year={2026}
}

@inproceedings{zhaolearning,
  title={Learning to Reason over Continuous Tokens with Reinforcement Learning},
  author={Zhao, Yiran and Xu, Yuhui and Sahoo, Doyen and Xiong, Caiming and Li, Junnan},
  booktitle={The Fourteenth International Conference on Learning Representations},
  year={2026}
}

@article{du2025latent,
  title={Latent thinking optimization: Your latent reasoning language model secretly encodes reward signals in its latent thoughts},
  author={Du, Hanwen and Dong, Yuxin and Ning, Xia},
  journal={arXiv preprint arXiv:2509.26314},
  year={2025}
}

@article{yue2025hybrid,
  title={Hybrid latent reasoning via reinforcement learning},
  author={Yue, Zhenrui and Jin, Bowen and Zeng, Huimin and Zhuang, Honglei and Qin, Zhen and Yoon, Jinsung and Shang, Lanyu and Han, Jiawei and Wang, Dong},
  journal={arXiv preprint arXiv:2505.18454},
  year={2025}
}

@article{gpt4,
  title={Gpt-4 technical report},
  author={Achiam, Josh and Adler, Steven and Agarwal, Sandhini and Ahmad, Lama and Akkaya, Ilge and Aleman, Florencia Leoni and Almeida, Diogo and Altenschmidt, Janko and Altman, Sam and Anadkat, Shyamal and others},
  journal={arXiv preprint arXiv:2303.08774},
  year={2023}
}

@article{gpt5,
  title={Openai gpt-5 system card},
  author={Singh, Aaditya and Fry, Adam and Perelman, Adam and Tart, Adam and Ganesh, Adi and El-Kishky, Ahmed and McLaughlin, Aidan and Low, Aiden and Ostrow, AJ and Ananthram, Akhila and others},
  journal={arXiv preprint arXiv:2601.03267},
  year={2025}
}

@article{llama3,
  title={The llama 3 herd of models},
  author={Grattafiori, Aaron and Dubey, Abhimanyu and Jauhri, Abhinav and Pandey, Abhinav and Kadian, Abhishek and Al-Dahle, Ahmad and Letman, Aiesha and Mathur, Akhil and Schelten, Alan and Vaughan, Alex and others},
  journal={arXiv preprint arXiv:2407.21783},
  year={2024}
}

@article{gemini,
  title={Gemini: a family of highly capable multimodal models},
  author={Team, Gemini and Anil, Rohan and Borgeaud, Sebastian and Alayrac, Jean-Baptiste and Yu, Jiahui and Soricut, Radu and Schalkwyk, Johan and Dai, Andrew M and Hauth, Anja and Millican, Katie and others},
  journal={arXiv preprint arXiv:2312.11805},
  year={2023}
}

@inproceedings{lightman2023let,
  title={Let's verify step by step},
  author={Lightman, Hunter and Kosaraju, Vineet and Burda, Yuri and Edwards, Harrison and Baker, Bowen and Lee, Teddy and Leike, Jan and Schulman, John and Sutskever, Ilya and Cobbe, Karl},
  booktitle={The twelfth international conference on learning representations},
  year={2023}
}

@article{li2022competition,
  title={Competition-level code generation with alphacode},
  author={Li, Yujia and Choi, David and Chung, Junyoung and Kushman, Nate and Schrittwieser, Julian and Leblond, R{\'e}mi and Eccles, Tom and Keeling, James and Gimeno, Felix and Dal Lago, Agustin and others},
  journal={Science},
  volume={378},
  number={6624},
  pages={1092--1097},
  year={2022},
  publisher={American Association for the Advancement of Science}
}

@article{nijkamp2022codegen,
  title={Codegen: An open large language model for code with multi-turn program synthesis},
  author={Nijkamp, Erik and Pang, Bo and Hayashi, Hiroaki and Tu, Lifu and Wang, Huan and Zhou, Yingbo and Savarese, Silvio and Xiong, Caiming},
  journal={arXiv preprint arXiv:2203.13474},
  year={2022}
}

@article{shinn2023reflexion,
  title={Reflexion: Language agents with verbal reinforcement learning},
  author={Shinn, Noah and Cassano, Federico and Gopinath, Ashwin and Narasimhan, Karthik and Yao, Shunyu},
  journal={Advances in neural information processing systems},
  volume={36},
  pages={8634--8652},
  year={2023}
}

@inproceedings{hao2023reasoning,
  title={Reasoning with language model is planning with world model},
  author={Hao, Shibo and Gu, Yi and Ma, Haodi and Hong, Joshua and Wang, Zhen and Wang, Daisy and Hu, Zhiting},
  booktitle={Proceedings of the 2023 Conference on Empirical Methods in Natural Language Processing},
  pages={8154--8173},
  year={2023}
}

@article{grpo,
  title={Deepseekmath: Pushing the limits of mathematical reasoning in open language models},
  author={Shao, Zhihong and Wang, Peiyi and Zhu, Qihao and Xu, Runxin and Song, Junxiao and Bi, Xiao and Zhang, Haowei and Zhang, Mingchuan and Li, YK and Wu, Yang and others},
  journal={arXiv preprint arXiv:2402.03300},
  year={2024}
}

@article{zhu2025flowrl,
  title={Flowrl: Matching reward distributions for llm reasoning},
  author={Zhu, Xuekai and Cheng, Daixuan and Zhang, Dinghuai and Li, Hengli and Zhang, Kaiyan and Jiang, Che and Sun, Youbang and Hua, Ermo and Zuo, Yuxin and Lv, Xingtai and others},
  journal={arXiv preprint arXiv:2509.15207},
  year={2025}
}

@article{r1,
  title={Deepseek-r1: Incentivizing reasoning capability in llms via reinforcement learning},
  author={Guo, Daya and Yang, Dejian and Zhang, Haowei and Song, Junxiao and Wang, Peiyi and Zhu, Qihao and Xu, Runxin and Zhang, Ruoyu and Ma, Shirong and Bi, Xiao and others},
  journal={arXiv preprint arXiv:2501.12948},
  year={2025}
}

@article{adamw,
  title={Decoupled weight decay regularization},
  author={Loshchilov, Ilya and Hutter, Frank},
  journal={arXiv preprint arXiv:1711.05101},
  year={2017}
}
\bibliographystyle{unsrtnat}

\newpage
\appendix
\section{Notation Definitions}
The major notations used in this paper are listed in Table~\ref{tab:notation}.

\begin{table}[H]
\caption{Major Notations.}
\vspace{1em}
\centering
\scalebox{0.93}{
\begin{tabular}{ll}
\toprule
\textbf{Notation} & \textbf{Definition} \\
\midrule
$x$ & Input question. \\
$y$ & Target answer. \\
$h_x$ & Embedding of the input $x$ produced by the backbone encoder, i.e., $h_x = p_\phi(x)$. \\
$\mathbf z_t$ & Continuous latent thought sampled at step $t$. \\
$s_t$ & Context state $(h_x, z_{1:t})$. \\
$T$ & Number of latent thoughts in $\tau$ \\
$\bot$ & Stop token indicating termination of the latent trajectory. \\
$\tau=(\mathbf z_{1:T},\bot)$ & Variable-length latent thought trajectory ending with stop token $\bot$. \\
$q_\varphi(\mathbf z_{t+1}\mid s_t)$ & Continuous latent thought transition density. \\
$p_\psi(<\mathrm{eos_r}>\mid s_t)$ & Stopping probability of latent reasoning at prefix state $s_t$. \\
$\pi^\bot(s_t)$ & Shorthand for $p_\psi(<\mathrm{eos_r}>\mid s_t)$. \\
$F(s_t)$ & Flow assigned to state $s_t$. \\
$\mathcal R_{x,y}(s_t\to\bot)$ & Terminal reward for stopping from state $s_t$ given the input $x$ and target $y$. \\
$\chi_{i:j}$ & Continuous SubTB residual over subtrajectories $s_i\to\cdots\to s_j$. \\
\bottomrule
\end{tabular}}
\label{tab:notation}
\end{table}

\section{Additional Implementation Details}
\label{app:imp}
In this section, we provide comprehensive details regarding baselines and training details.

\subsection{Baseline Details}
\label{app:baseline}
For the main experiments, we employ various explicit and latent-based reasoning methods as baselines. The reported metric reasoning length ($\# L$) measures the number of reasoning steps before producing the final answer: for explicit reasoning methods, this corresponds to the number of explicit reasoning tokens; for latent reasoning approaches, it corresponds to the number of latent reasoning steps; and for implicit one-pass methods such as iCoT, $\# L$ is defined as $0$. Importantly, latent reasoning methods typically implement each latent step with approximately the same forward computation as one explicit decoding step, resulting in comparable per-step computational overhead in different reasoning paradigms. Thus, $\# L$ serves as a practical metric for evaluating reasoning efficiency across both explicit and latent reasoning frameworks. The detailed description of the above baselines is provided below:

\textbf{CoT}~\citep{cot}: This baseline uses chain-of-thought prompting templates to elicit explicit intermediate reasoning steps before producing the final answer.

\textbf{Assist-CoT}~\citep{softcot}: This baseline uses a smaller assistant model to generate explicit reasoning tokens, which are then provided to the LLM as chain-of-thought prompts. The reasoning length ($\# L$) includes both soft thought tokens and intermediate reasoning steps.

\textbf{SoftCoT++}~\citep{softcot2}: This method generates multiple diverse soft thought representations through specialized initial tokens, and further encourages diversity among them with a contrastive learning objective. These soft thoughts are then used to guide the LLM's explicit reasoning process. The reasoning length ($\# L$) includes both soft thought tokens and intermediate reasoning steps.

\textbf{iCoT}~\citep{icot}: This baseline progressively drops intermediate reasoning steps during fine-tuning, encouraging the model to internalize the reasoning process while preserving task performance.

\textbf{CODI}~\citep{codi}: This baseline uses self-distillation to align the hidden activations of latent thoughts with CoT trajectories, thereby transferring explicit reasoning patterns into the latent space.

\textbf{Coconut}~\citep{coconut}: This baseline recursively feeds the last hidden state of the LLM back as a latent thought embedding, using it as the next-step input to support subsequent reasoning.

\textbf{CoLaR}~\citep{colar}: This baseline dynamically compresses reasoning chains into latent variables, where each latent state summarizes multiple consecutive reasoning tokens. It trains a latent head to autoregressively predict compressed embeddings and uses reinforcement learning to encourage diverse and concise latent reasoning paths.

\textbf{ReGuLaR}~\citep{regular}: This baseline formulates latent reasoning within a variational framework, where latent reasoning states are sampled from posterior distributions conditioned on previous states. It renders explicit CoT segments as images and uses their visual representations to regularize the latent states.

For ablation studies, we employ RL- and GFlowNet-based methods as baselines. The detailed description of the above baselines is provided below:

\textbf{GRPO}~\citep{grpo}: This baseline samples a group of outputs 
$\{o_1,o_2,\ldots,o_G\}$ from the old policy $p_{\Theta_{\mathrm{old}}}$, 
where $G$ denotes the group size. Each output $o_i$ consists of a latent 
reasoning trajectory $\tau_i$ and a final answer. GRPO then optimizes the 
current policy $p_{\Theta}$ by minimizing the following clipped objective:
\begin{equation}
\mathcal{L}_{\mathrm{GRPO}}
=
-\frac{1}{G}\sum_{i=1}^{G}
\min\left(
\frac{p_{\Theta}(o_i \mid x)}{p_{\Theta_{\mathrm{old}}}(o_i \mid x)} A_i,
\operatorname{clip}\left(
\frac{p_{\Theta}(o_i \mid x)}{p_{\Theta_{\mathrm{old}}}(o_i \mid x)},
1-\epsilon, 1+\epsilon
\right) A_i
\right).
\end{equation}
Here, $A_i$ is computed as a group-normalized reward:
\begin{equation}
A_i =
\frac{r_i - \operatorname{mean}(r_1,r_2,\ldots,r_G)}
{\operatorname{std}(r_1,r_2,\ldots,r_G)}.
\end{equation}
We use $\mathcal{R}_{x,y}(\tau_i)$ as the reward, i.e.,
\begin{equation}
r_i = \mathcal{R}_{x,y}(\tau_i).
\end{equation}
We replace the LTF objective Eq.~\eqref{eq:overall_loss_ltf} with $\mathcal{L}_{\mathrm{GRPO}}$ 
and keep other training configurations same.

\textbf{Detailed Balance (DB)}~\citep{gflownet}: This variant uses the same overall objective as LTF 
(Eq.~\eqref{eq:overall_loss_ltf}), but replaces 
$\mathcal{L}_{\mathrm{flow}}$ with a Detailed Balance loss. 
It can be viewed as a one-step special case of Sub-Trajectory Balance, 
where flow consistency is enforced only between adjacent states. 
Specifically,
\begin{equation}
\label{eq:continuous_db_loss}
    \mathcal{L}_{\mathrm{DB}}
    =
    \mathbb{E}_{\tau\sim q_\varphi(\cdot\mid s_0)}
    \left[
    \sum_{t=0}^{T-1}
    \chi_{t:t+1}^{2}
    \right],
\end{equation}
where $\chi_{t:t+1}$ is defined in Eq.~\eqref{eq:continuous_subtb_residual}.
All other training configurations are kept unchanged.

\textbf{Trajectory Balance (TB)}~\citep{tb}: This variant uses the same overall objective as LTF 
(Eq.~\eqref{eq:overall_loss_ltf}), but replaces 
$\mathcal{L}_{\mathrm{flow}}$ with a Trajectory Balance loss. 
Unlike DB, which enforces local one-step consistency, TB enforces flow 
consistency only over the complete sampled trajectory. Specifically,
\begin{equation}
\label{eq:continuous_tb_loss}
    \mathcal{L}_{\mathrm{TB}}
    =
    \mathbb{E}_{\tau\sim q_\varphi(\cdot\mid s_0)}
    \left[
    \chi_{0:T}^{2}
    \right],
\end{equation}
where $\chi_{0:T}$ is given by 
Eq.~\eqref{eq:continuous_subtb_residual}. 
All other training configurations are kept unchanged.

\subsection{Training Details}
For all experiments, the Latent Head in LTF is implemented as a three-layer MLP, 
with hidden dimensions matching the hidden size of the LLM backbone. We set $\texttt{rollout\_n}=20$ and $T_{\max}=32$ during training, meaning that the model samples $20$ reasoning trajectories with maximum latent reasoning length of $32$ for each prompt. For transfer learning settings, we use the models finetuned on GSM8K-Aug to study the out-of-domain generalization performance. In the ablation study on latent-thought exploration objectives, we use the same 
setting, i.e., $\texttt{rollout\_n}=20$ and $T_{\max}=32$, for all compared methods, including GRPO, DB, and TB, to ensure a fair comparison. Following prior work~\citep{regular}, 
we construct the reference representation $r$ as the embedding of golden explicit rationale in the training dataset. All hyperparameters in LTF were selected by grid search on the validation set and then fixed across all experiments. We set $\lambda_c=0.03$ to control the cost penalty in the terminal reward. In validation experiments, LTF was stable for $\lambda_c \in (0.01, 0.04)$, and we therefore use the value $0.03$ as the default setting. We set $\lambda_{\mathrm{ans}}=1.0$ for the answer-generation loss, so that the answer objective remains on the same scale as the main training objective. For the prior loss, we use a piecewise-linear annealing schedule for $\lambda_{\mathrm{prior}}$. Specifically, $\lambda_{\mathrm{prior}}$ is initialized to $3.0$ and linearly decayed to $0.1$ over 100 epochs, with the value updated every five epochs. This schedule imposes a strong prior constraint at the beginning of training and gradually relaxes the constraint as training proceeds, allowing the reward-driven objective to have a larger influence in later stages. We found that LTF is generally robust to small variations around these settings. 

\section{Extended Experimental Results}
\label{app:exp}
\subsection{Extended Empirical Results}
\label{app:ex}
\begin{table*}[t!]
\centering
\caption{Performance comparison under finetuning setting (GSM8K-Aug) and transfer learning setting (GSM-Hard, SVAMP, MultiArith) using LLaMA-3.2-1B-Instruct. We report the averaged number with 95\% confidence interval on Accuracy (Acc. \%) and Reasoning Length ($\#$ L).}
\label{tab:app}
\setlength{\tabcolsep}{3pt}
\scalebox{0.82}{
\begin{tabular}{c||cccccccc}
\toprule
\multirow{2}{*}[-0.55ex]{\textbf{Method}} & \multicolumn{2}{c}{\textbf{GSM8K-Aug}} & \multicolumn{2}{c}{\textbf{GSM-Hard}} & \multicolumn{2}{c}{\textbf{SVAMP}} & \multicolumn{2}{c}{\textbf{MultiArith}}\\
\cmidrule(lr){2-3} \cmidrule(lr){4-5} \cmidrule(lr){6-7} \cmidrule(lr){8-9}
& \multicolumn{1}{c}{Acc.($\uparrow$)} & \multicolumn{1}{c}{\#~L($\downarrow$)} & \multicolumn{1}{c}{Acc.($\uparrow$)} & \multicolumn{1}{c}{\#~L($\downarrow$)} & \multicolumn{1}{c}{Acc.($\uparrow$)} & \multicolumn{1}{c}{\#~L($\downarrow$)} & \multicolumn{1}{c}{Acc.($\uparrow$)} & \multicolumn{1}{c}{\#~L($\downarrow$)} \\
\midrule
CoT & $15.70_{\pm0.18}$ & $120.37_{\pm0.17}$ & $2.74_{\pm0.12}$ & $120.46_{\pm0.23}$ & $27.17_{\pm0.23}$ & $120.80_{\pm0.18}$ & $28.29_{\pm0.31}$ & $120.14_{\pm0.19}$ \\
Assist-CoT & $16.30_{\pm0.19}$ & $119.85_{\pm0.19}$ & $3.02_{\pm0.15}$ & $125.19_{\pm0.24}$ & $28.04_{\pm0.29}$ & $123.02_{\pm0.18}$ & $30.06_{\pm0.34}$ & $121.86_{\pm0.20}$ \\
SoftCoT++ & $17.04_{\pm0.21}$ & $119.73_{\pm0.20}$ & $3.77_{\pm0.16}$ & $123.56_{\pm0.26}$ & $30.96_{\pm0.31}$ & $123.42_{\pm0.25}$ & $32.13_{\pm0.36}$ & $120.66_{\pm0.23}$ \\
iCoT & $19.80_{\pm0.22}$ & $0.00_{\pm0.00}$ & $3.82_{\pm0.16}$ & $0.00_{\pm0.00}$ & $36.42_{\pm0.52}$ & $0.00_{\pm0.00}$ & $38.19_{\pm0.64}$ & $0.00_{\pm0.00}$ \\
CODI & $13.30_{\pm0.64}$ & $6.00_{\pm0.00}$ & $2.94_{\pm0.24}$ & $6.00_{\pm0.00}$ & $21.65_{\pm0.70}$ & $6.00_{\pm0.00}$ & $19.20_{\pm0.82}$ & $6.00_{\pm0.00}$ \\
Coconut & $20.50_{\pm0.74}$ & $6.00_{\pm0.00}$ & $4.87_{\pm0.31}$ & $6.00_{\pm0.00}$ & $38.90_{\pm0.71}$ & $6.00_{\pm0.00}$ & $41.30_{\pm0.67}$ & $6.00_{\pm0.00}$ \\
CoLaR & $26.60_{\pm0.17}$ & $5.63_{\pm0.05}$ & $6.22_{\pm0.14}$ & $7.03_{\pm0.06}$ & $47.22_{\pm0.30}$ & $2.94_{\pm0.03}$ & $86.93_{\pm0.21}$ & $3.22_{\pm0.05}$ \\
ReGuLaR & $34.58_{\pm0.22}$ & $3.69_{\pm0.21}$ & $8.25_{\pm0.14}$ & $4.10_{\pm0.43}$ & $48.19_{\pm0.39}$ & $2.10_{\pm0.19}$ & $88.97_{\pm0.27}$ & $2.29_{\pm0.28}$ \\
\rowcolor{highlightred}
LTF & $\textbf{37.09}_{\pm0.19}$ & $\textbf{3.34}_{\pm0.09}$ & $\textbf{8.71}_{\pm0.13}$ & $\textbf{3.97}_{\pm0.21}$ & $\textbf{52.01}_{\pm0.27}$ & $\textbf{2.11}_{\pm0.06}$ & $\textbf{90.37}_{\pm0.19}$ & $\textbf{2.17}_{\pm0.08}$ \\
\bottomrule
\end{tabular}}
\end{table*}
\begin{table*}[t!]
\centering
\caption{Ablation on reference-prior regularization (RPR) using the LLaMA-3.1 Instruct 1B backbone. We report the averaged Accuracy (Acc. \%) and Reasoning Length ($\#$ L).}
\label{tab:app1}
\setlength{\tabcolsep}{8pt}
\small{
\begin{tabular}{l|cccccccc}
\toprule
\multirow{2}{*}[-0.55ex]{\textbf{Method}} 
& \multicolumn{2}{c}{\textbf{GSM8K-Aug}} 
& \multicolumn{2}{c}{\textbf{ASDiv-Aug}} 
& \multicolumn{2}{c}{\textbf{DU}} 
& \multicolumn{2}{c}{\textbf{Average}}\\
\cmidrule(lr){2-3} \cmidrule(lr){4-5} \cmidrule(lr){6-7} \cmidrule(lr){8-9}
& Acc.($\uparrow$) & \#~L($\downarrow$) 
& Acc.($\uparrow$) & \#~L($\downarrow$) 
& Acc.($\uparrow$) & \#~L($\downarrow$) 
& Acc.($\uparrow$) & \#~L($\downarrow$) \\
\midrule
LTF & 37.09 & \textbf{3.34} & \textbf{75.11} & \textbf{1.22} & \textbf{66.83} & \textbf{1.18} & \textbf{59.68} & \textbf{1.91} \\
\textcolor{gray}{\hspace{0.8em}w/o RPR} 
& \textbf{37.21}	&3.72	&75.04	&2.38	&65.70	&2.14&59.32	&2.75 \\
\bottomrule
\end{tabular}}
\end{table*}
\begin{table*}[t!]
\centering
\caption{Ablation on number of latent thought chains $N$ during test time using the LLaMA-3.1 Instruct 1B backbone. We report the averaged Accuracy (Acc. \%) and Reasoning Length ($\#$ L).}
\label{tab:test}
\setlength{\tabcolsep}{10.8pt}
\scalebox{0.9}{
\begin{tabular}{c|cccccccc}
\toprule
\multirow{2}{*}[-0.55ex]{$N$} & \multicolumn{2}{c}{\textbf{GSM8K-Aug}} & \multicolumn{2}{c}{\textbf{ASDiv-Aug}} & \multicolumn{2}{c}{\textbf{DU}} & \multicolumn{2}{c}{\textbf{Average}}\\
\cmidrule(lr){2-3} \cmidrule(lr){4-5} \cmidrule(lr){6-7} \cmidrule(lr){8-9}
& Acc.($\uparrow$) & \#~L($\downarrow$) & Acc.($\uparrow$) & \#~L($\downarrow$) & Acc.($\uparrow$) & \#~L($\downarrow$) & Acc.($\uparrow$) & \#~L($\downarrow$) \\
\midrule
1 & 37.09 & 3.34 & 75.11 & 1.22 & 66.83 & 1.18 & 59.68 & 1.91 \\
5 & 38.13 & 3.38 & 77.62 & 1.24 & 68.03 & 1.20 & 61.26 & 1.94 \\
10 & 38.72 & 3.37 & 78.73 & 1.23 & 68.94 & 1.20 & 62.13 & 1.93 \\
\bottomrule
\end{tabular}}
\end{table*}

\noindent{\textbf{Main Results with 95\% Confidence Interval.}} Table~\ref{tab:app} shows finetuning and transfer learning performance comparison on reasoning datasets under LLaMA-3.2-1B-Instruct with 95\% confidence interval ($\pm$) on Accuracy (Acc. \%) and Reasoning Length ($\#$ L). These results indicate that LTF achieves better accuracy--efficiency tradeoffs across tasks. This observation can be attributed to the Entropy-Weighted SubTB objective, which encourages compact and informative latent states rather than redundant reasoning trajectories, highlighting the stability and broad applicability of LTF across model families and scales.

\noindent{\textbf{Ablation Analysis on Reference-Prior Regularization (RPR).}}
Table~\ref{tab:app1} ablates the effect of reference-prior regularization (RPR) (Sec.~\ref{sec:reference_prior}) in LTF.
Compared with LTF w/o RPR, LTF improves the average accuracy from $59.32\%$ to $59.68\%$, while reducing the average latent reasoning length from $2.75$ to $1.91$. Notably, LTF w/o RPR slightly outperforms LTF on GSM8K-Aug ($37.21$ vs. $37.09$), highlighting the difficulty of latent-space exploration, where optimization can become trapped in local optima. However, LTF achieves better overall performance while producing more concise reasoning paths, validating the effectiveness of RPR in stabilizing exploration.

\noindent{\textbf{Analysis on Test-Time Scaling.}}
Table~\ref{tab:test} evaluates how the number of latent thought chains $N$ affects inference performance. Increasing $N$ consistently improves accuracy across all three benchmarks, while leaving the average reasoning length nearly unchanged. Notably, raising $N$ from $1$ to $10$ improves average accuracy from $59.68\%$ to $62.13\%$ ($+2.45$), whereas the average reasoning length changes only marginally from $1.91$ to $1.93$. These results demonstrate a key advantage of LTF: it supports effective test-time scaling by sampling multiple compact latent reasoning trajectories, rather than relying on substantially longer reasoning traces. We further observe that LTF can adaptively allocate different numbers of reasoning steps to problems of varying difficulty.

This trend aligns with the intended Monte Carlo inference mechanism of LTF. By sampling more latent thought chains, the model increases the probability of identifying a high-reward reasoning trajectory. At the same time, the diminishing improvement from $N=5$ to $N=10$ suggests that a modest number of samples is sufficient to capture most of the attainable benefit. Overall, these results provide empirical evidence that the learned reward-induced sampling posterior distribution in LTF is effective for improving inference-time performance efficiently.

\subsection{Analysis of Reasoning Trajectories}
\label{app:ex1}
\begin{table*}[t!]
\centering
\caption{
Average reasoning entropy with $95\%$ confidence interval of latent reasoning trajectories across different reasoning methods using the LLaMA-3.1 Instruct 8B backbone. Average reasoning entropy $\Xi(\tau)$ is calculated via Eq.~\eqref{eq:entropy}.
}
\label{tab:reasoning_entropy}
\setlength{\tabcolsep}{50pt}
\scalebox{0.95}{
\begin{tabular}{c|c}
\toprule
\textbf{Method} 
& \textbf{Average Reasoning Entropy $\Xi(\tau)$} \\
\midrule
CoLaR 
& $0.013\pm0.009$ \\
ReGuLaR 
& $0.019\pm0.002$ \\
LTF \textcolor{gray}{w/o EW}
& $0.030\pm0.006$ \\
LTF
& $0.024\pm0.003$ \\
\bottomrule
\end{tabular}}
\end{table*}
To better understand the latent reasoning dynamics learned by LTF, we analyze the average reasoning entropy of latent reasoning paths generated by CoLaR, ReGuLaR, LTF w/o entropy weighting (EW), and LTF under LLaMA-3.1 Instruct 8B backbone. Entropy quantifies how much information content the latent reasoning trajectory carries. A higher entropy often indicates a richer spread of information across many dimensions, reflecting diverse, less redundant features and better information preservation. Conversely, a lower entropy usually reflects concentrated eigenvalue spectra, suggesting that the latent representations may contain redundant information. Given a latent reasoning trajectory $\tau=(s_0,\mathbf{z}_{1:T})$, we define its \textit{average reasoning entropy} as
\begin{equation}
\label{eq:entropy}
    \Xi(\tau)
    =
    \frac{1}{T}
    \sum_{t=0}^{T-1}
    \mathcal H
    \left[
        q_\varphi(\mathbf z_{t+1}\mid s_t)
    \right],
\end{equation}
where $\mathcal{H}(\cdot)$ denotes differential entropy.

Our empirical study investigates the property of  latent reasoning paths based on test samples drawn from fine-tuning tasks. As shown in Table~\ref{tab:reasoning_entropy}, CoLaR exhibits the lowest reasoning entropy (0.013), suggesting that its latent reasoning trajectories are highly deterministic and potentially prone to trajectory collapse. ReGuLaR yields higher entropy (0.019), indicating that it encourages more stochastic latent reasoning behaviors. LTF w/o entropy weighting further increases the entropy (0.030), producing even more stochastic but less structured reasoning trajectories.

In contrast, LTF achieves lower entropy (0.024) than LTF w/o entropy weighting while still maintaining higher stochasticity than CoLaR and ReGuLaR. This pattern suggests that entropy-weighted sub-trajectories do not simply maximize stochasticity; rather, they regulate the degree of exploration in latent reasoning. By adaptively allocating training credit to sub-trajectories with appropriate uncertainty, LTF encourages reasoning paths that are both diverse and structured, which helps explain its stronger downstream performance.

\textbf{These observations indicate the existence of an effective entropy regime for latent reasoning and we denote the threshold as \textit{effective entropy threshold}.} When reasoning entropy is too low and under the effective entropy threshold, latent trajectories may collapse into overly deterministic patterns, limiting information contained in the latent reasoning paths and weakening generalization. Moderate increases in entropy can improve reasoning performance by encouraging diverse latent trajectories. However, once entropy becomes excessive beyond the effective entropy threshold, the reasoning process may become overly stochastic, making the learned trajectories less reliable and less effecitve for downstream tasks. Thus, identifying the effective entropy regime is crucial for regulating latent reasoning dynamics and improving reasoning performance.

\section{Broader Impact}
The large language models employed in this research may reflect biases or generate sensitive or potentially offensive responses, intended solely for academic and scientific purposes. The opinions expressed within generated outputs do not represent the views of the authors. We remain committed to fostering the development of AI technologies that align with ethical standards and reflect societal values.



\end{document}